\documentclass[letterpaper]{article} 
\usepackage[]{aaai23}  
\usepackage{times}  
\usepackage{helvet}  
\usepackage{courier}  
\usepackage[hyphens]{url}  
\usepackage{graphicx} 
\usepackage{natbib}  
\usepackage{caption} 
\usepackage{algorithm}
\usepackage{algorithmic}

\usepackage{newfloat}
\usepackage{listings}
\DeclareCaptionStyle{ruled}{labelfont=normalfont,labelsep=colon,strut=off} 
\floatstyle{ruled}
\newfloat{listing}{tb}{lst}{}
\floatname{listing}{Listing}
\usepackage{booktabs}
\usepackage{subfigure}
\usepackage{amsmath}
\usepackage{amssymb}
\usepackage{multirow}
\usepackage[table,xcdraw]{xcolor}         

\definecolor{sp1}{HTML}{798EAC}
\definecolor{sp2}{HTML}{83B4A8}
\definecolor{sp3}{HTML}{A2AA71}
\definecolor{sp4}{HTML}{C5AA71}
\definecolor{sp5}{HTML}{C68D7D}

\title{Variational Transformer: A Framework Beyond the Trade-off between Accuracy and Diversity for Image Captioning}
\author{
    Longzhen Yang,
    Yihang Liu,
    Yitao Peng,
    Lianghua He \equalcontrib
}
\affiliations{
    \textsuperscript{\rm 1}Association for the Advancement of Artificial Intelligence\\

    College of Electronic and Information Engineering, Tongji University\\
    4800 Cao'an Highway, Shanghai, China 201804 \\
    \texttt{\{yanglongzhen, 2111131, 2111132, helianghua\}@tongji.edu.cn}
}

\begin{document}

\maketitle

\begin{abstract}
Accuracy and Diversity are two essential metrizable manifestations in generating natural and semantically correct captions. Many efforts have been made to enhance one of them with another decayed due to the trade-off gap. In this work, we will show that the inferior standard of accuracy draws from human annotations (leave-one-out) are not appropriate for machine-generated captions. To improve diversity with a solid accuracy performance, we exploited a novel Variational Transformer framework. By introducing the "Invisible Information Prior" and the "Auto-selectable GMM", we instruct the encoder to learn the precise language information and object relation in different scenes for accuracy assurance. By introducing the "Range-Median Reward" baseline, we retain more diverse candidates with higher rewards during the RL-based training process for diversity assurance. Experiments show that our method achieves the simultaneous promotion of accuracy (CIDEr) and diversity (self-CIDEr), up to $\mathbf{1.1}$ and $\mathbf{4.8}$ percent. Also, our method got the most similar performance of the semantic retrieval compared to human annotations, with $\mathbf{50.3}$ ($50.6$ of human) for R@1(i2t).
\end{abstract}

\section{Introduction}
\label{intro}

\begin{figure}[]
    \centering
    \includegraphics[width=0.47\textwidth]{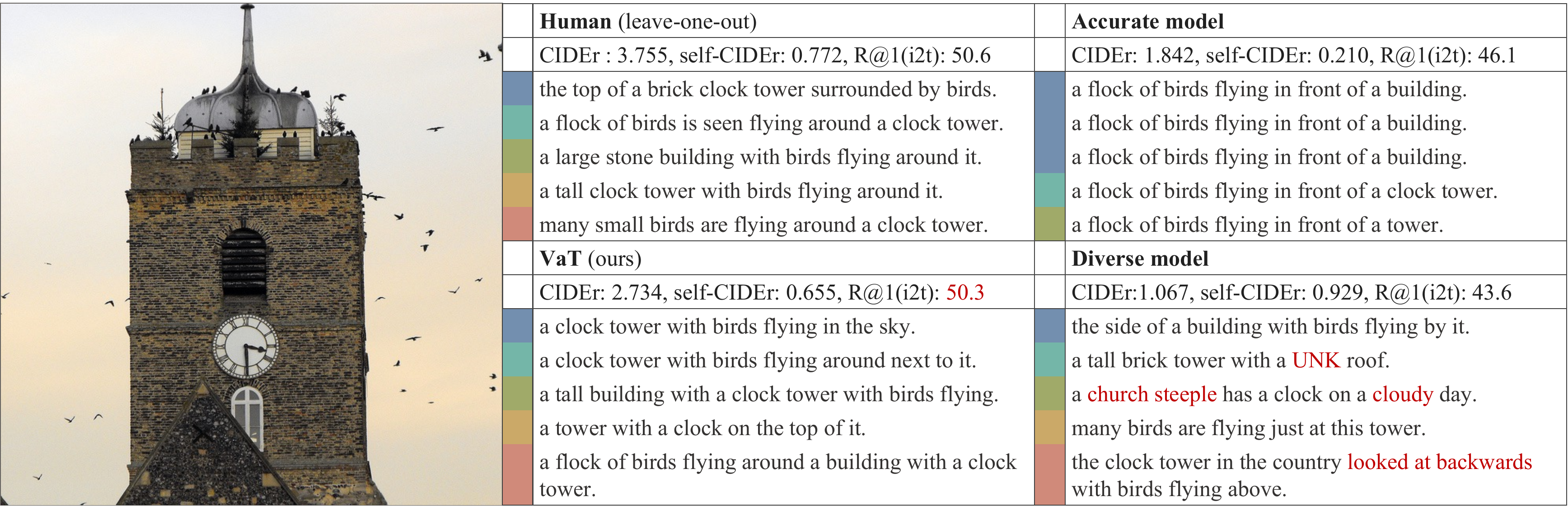}
    \caption{Captions generated from different models. Diverse captions with low accuracy scores can be deceptive. 1) We use CIDEr and self-CIDEr to present the performance of accuracy and diversity, respectively. 2) We denote distinct captions with different colors. 3) R@1 is the rate of a correctly retrieved groundtruth given top 1 candidate.}
    \label{fig:introduction}
\end{figure}

Generating diverse and accurate captions is a challenging task. Though, recent method in \cite{shi2021off-policy} did achieve close numerical results to human ground truths (leave-one-out~\cite{Wang2019Describing}) in both accuracy and diversity, it is still hard for Machine Learning (ML) models to reveal the true semantic performance with low accuracy, as shown in Figure~\ref{fig:introduction}. Common diversity metrics involve no accuracy measure~\cite{Luo2020Analysis}, thus, can be misleading with wrong words to manifest inflating diversity scores. From this observation, we suppose that good diverse generations should be established on a solid accuracy performance. 

\begin{figure}[t]
    \centering
    \includegraphics[width=0.46\textwidth]{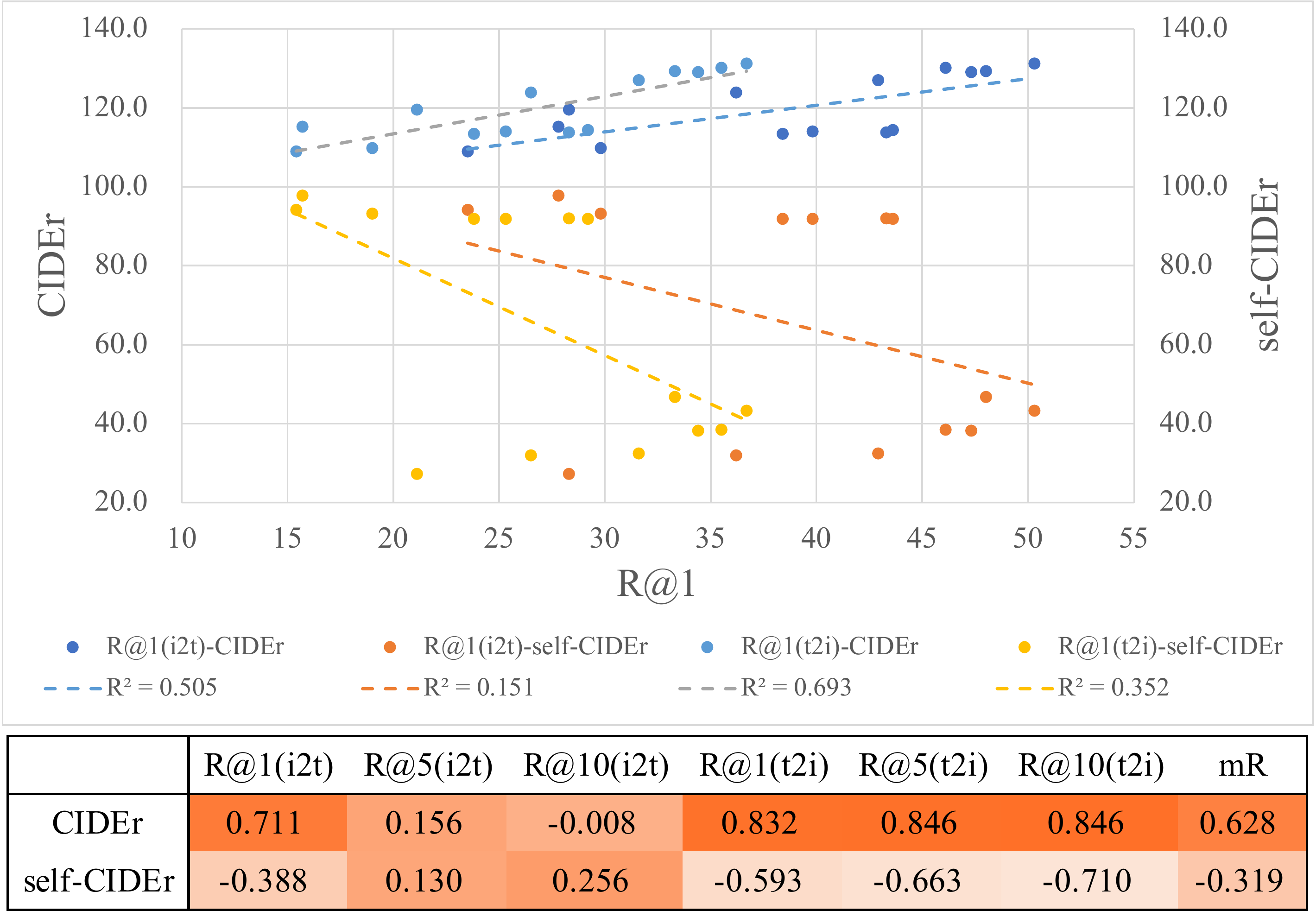}
    \caption{Semantic correlation analysis for accuracy and diversity metrics.}
    \label{fig:sca}
\end{figure}

To confirm this assumption, we need an objective measurement. However, as reported by~\cite{yamshchikov2021style}, there is still no metric, in current, that could distinguish
paraphrases form style transfers definitively. Another common way of evaluate the semantic similarity is image-text retrieval~\cite{frome2013devise, socher2014grounded, Mao2015Deep}, which utilizes the output probabilities of each model to construct the retrieval distribution. We, instead, chose to use a pre-trained retrieval model \cite{wang2020consensus} to make the evaluation process consistent. We tested on both image-to-text (i2t) and text-to-image (t2i) retrieval tasks. Results are consistent with our assumption, as shown in Figure~\ref{fig:introduction} and Figure~\ref{fig:sca}, only captions with solid high accuracy scores are semantically related with the target images.

From this point, we propose a novel Variational Transformer (VaT) framework with both accuracy and diversity assurance programs. In specific, we first design an "Invisible Information Prior" (IIP) using unmasked input sentences to navigate the posterior encoder to learn the precise language attention map. Then we modify the single gaussian prior of VAE into an "Auto-selectable GMM" (AGMM) to fit the complex distribution of object relations in different scenes. IIP and AGMM form together to construct our assurance program for accuracy. Second, we propose a reformulation of the self-critical sequence training (SCST)~\cite{Rennie2017Self-Critical} employing a "Rang-Median Reward" (RMR) baseline to retain more diverse candidates with higher rewards during the RL training. Our Variational framework and RMR form together to construct the assurance program for diversity.

In this work, our main contributions are: \textbf{1)} We uncover the relation between common metrics and the semantic correlation for image captioning and make a comprehensive analysis. \textbf{2)} We propose a novel framework to promote accuracy and diversity at the same time, with $\mathbf{1.1}$ and $\mathbf{4.8}$ percent boost respectively. \textbf{3)} We also achieve the best retrieval and trade-off performance in a newly proposed measurement, comparing with the human baseline (only 0.46 percent backward to the human-oriented boundary).

\section{Related works}

\paragraph{Image captioning.}

The most fundamental work in image captioning adopted the CNN-RNN-based Auto-Encoder (AE) structure as their backbone, including M-RNN~\cite{Mao2015Deep}, “Show and Tell”~\cite{Vinyals2015Show}, and Deep Visual-Semantic Alignments~\cite{Karpathy2015Deep}. Many follow-up efforts improve it with other technologies, like Attention Mechanism and Reinforcement Learning~\cite{Sutton2018Reinforcement}. For example, \cite{Rennie2017Self-Critical,Anderson2018Bottom-Up,Huang2019Attention,Cornia2020Meshed-Memory,ji2021mac} applied different attention structures to simulate the human attention on both vision and language area; \cite{Ranzato2015Sequence,Luo2018Discriminability,Bujimalla2020B-SCST:,nie2021trrl} utilized Reinforcement Learning algorithms to optimize the non-differentiable metrics, like CIDEr, directly on image captioning model; recently, \cite{Chen2020Say,Yang2019Auto-encoding} introduced an external scene graph structure based on the human intuition when looking at a brief description to augment with the potential related attributes and objects, and even made it controllable to say as you wish.

\paragraph{Improve accuracy with Attention and RL}

Attention Mechanism is one of the most influential techniques to strengthen the accuracy performance. One representative is the “Bottom-Up and Top-Down” mechanism~\cite{Anderson2018Bottom-Up}, which combined the attention mechanism in both vision and language area. As a vital derivative from Attention Mechanism, Transformer~\cite{Vaswani2017Attention} plays an important role in image captioning in recent time, as it dramatically improves the accuracy performance. \cite{Cornia2020Meshed-Memory,Huang2019Attention,ji2021mac,yang2021apn} introduced several direct or related improvements to the basic Transformer model, while the purposes are similar, to intensify the ability of information filtration and multi-modal capacity.

Reinforce Learning is another important method to improve accuracy dramatically. \cite{Rennie2017Self-Critical} proposed the classic SCST strategy using greedy sampled sentences to refine the outputs with less diversity but better accuracy. Many works made modification based on this method, where the critical point lies in the baseline of RL training. \cite{Luo2020Better, Bujimalla2020B-SCST:, nie2021trrl} proposed three different and effective variants of RL baseline. We compared them with our RMR baseline to show our strength of improving diversity without damaging the accuracy performance. 

\paragraph{Improve diversity with VAE.}

Variational Auto-Encoder~\cite{Kingma2014Auto-Encoding} is wildly used in generative tasks, as well as several variants like CVAE~\cite{Kingma2014Semi-supervised} and \cite{Sohn2015Learning} and beta-CVAE~\cite{Higgins2017beta-VAE}. In image captioning, \cite{Wang2017Diverse} proposed GMM-CVAE to estimate the KL-divergence between gaussian mixture distributions, using the extra object information from the detection model. In this work, we devised a similar AGMM variant that is trainable end-to-end and can better recognize object relations in different scenes. \cite{Chen2019Variational} is another exploration of Variational structure, which proposed a novel variational multi-modal inferring tree (similar to the syntax tree) to improve the lexical and syntactic diversity in captioning. At last, \cite{Luo2020Analysis,shi2021off-policy} are two works which also concerned about the relation between accuracy and diversity like the main purpose of this work. \cite{shi2021off-policy} proposed an off-policy strategy to increase the range of samples during RL training, which improves the diversity dramatically, however, also causes the same dramatic decrease of accuracy. In our work, we fit this problem successfully through a well-designed framework.

\paragraph{Semantic retrieval for image captioning.}

Semantic retrieval between image and text has been wildly used for evaluating the semantic similarity of images and generated captions in image captioning~\cite{frome2013devise, socher2014grounded, Mao2015Deep}. In this work, we adopted CVSE~\cite{wang2020consensus} to analyse the capability of accuracy and diversity metrics that can reveal the true semantic correlation between images and captions. Thanks to their extraordinary work on vision-language retrieval~\footnote{https://github.com/BruceW91/CVSE\label{cvsegitlink}}, we can easily and consistently evaluate the captions generated by different models. 

\begin{figure*}
\begin{center}
\includegraphics[width=0.95\linewidth]{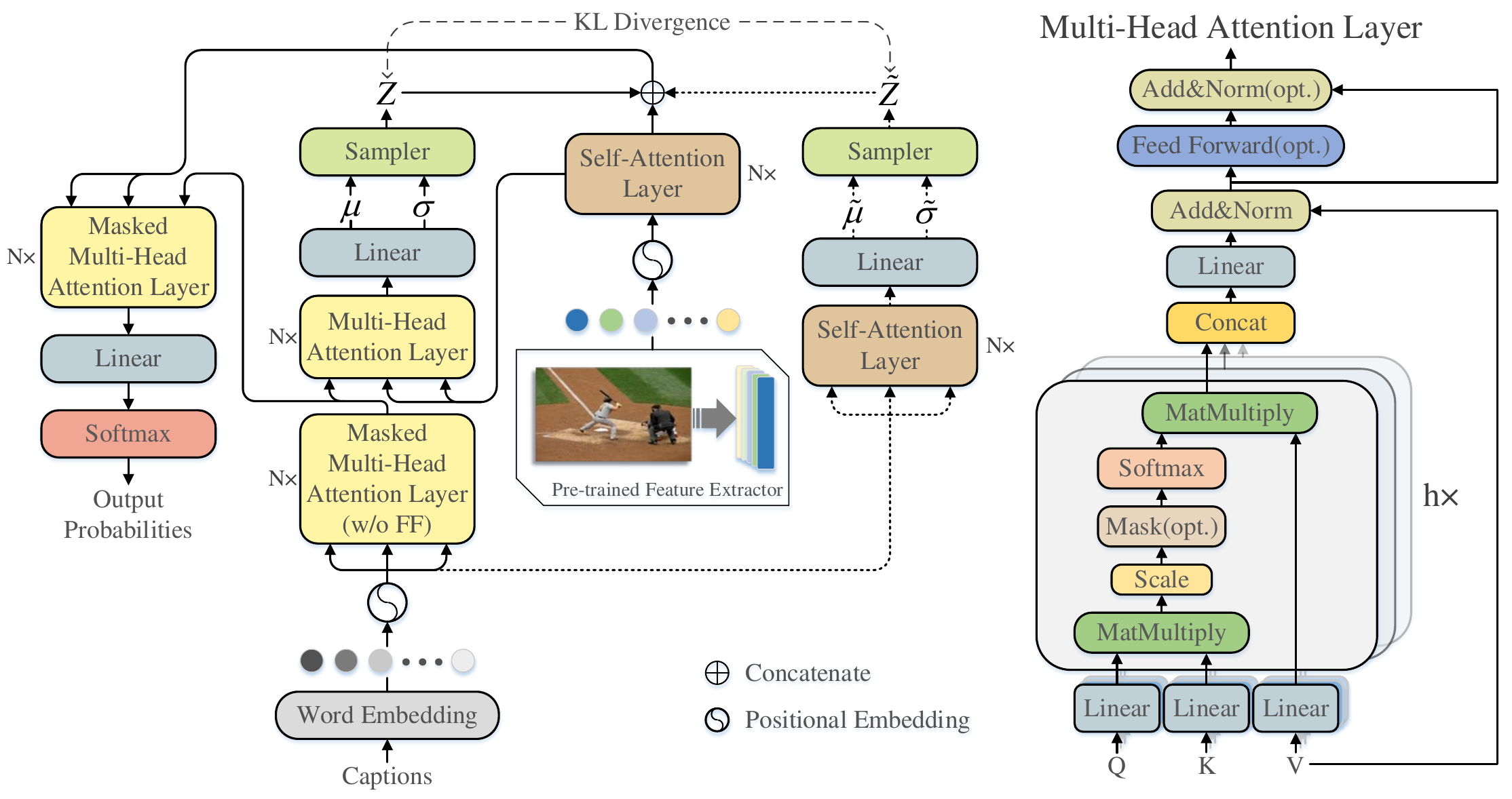}
\end{center}
   \caption{Overview of the proposed Variational Transformer architecture. The left figure shows the overall structure of our VaT model. The solid lines with the arrow present the inference route, while the dashed lines present the extra variational route at the training period. The right figure shows the specific Multi-Head Attention Layer in our model in contrast with the original Transformer Multi-Head Attention structure.}
\label{fig1}
\end{figure*}
\section{Semantic Correlation Analysis}

To verify the proposed assumption, we designed a simple test, which replaces human annotations with captions genereated by different models then runs the retrieval process through a fixed pre-trained model. In this work, we tested different models including those with high accuracy but low diversity scores and those with high diversity but low accuracy. 

In Figure~\ref{fig:sca}, we use R@$K$($K=1, 5, 10$) and mean recall (mR) to present the semantic correlation between images and generated captions. R@$K$ is the rate of a correctly retrieved groundtruth given top $K$ candidates. We draw the points of (R@1, CIDEr) and (R@1, self-CIDEr) pairs and their trendlines. $R^2$ presents the correlation between the trendlines and the points, higher is more matched. We also present the Pearson correlation heat map between metrics of captioning task and all the common recall value of retrieval task. The complete experimental results can be found in Appendix.

Through analysing the results of this section, we found three conclusions as follow.

1) The accuracy performance of generated captions has positive linear correlation with the semantic correlation of image-caption pairs, while the diversity does not.

2) We found that, compared with human groundtruth ($87.8$ on CIDEr, $88.6$ on self-CIDEr and $50.6$ on R@1-i2t), models that have both higher numerical scores on accuracy and diversity (e.g. $114.0$ on CIDEr and $89.8$ on self-CIDEr) achieved no close performance on retrieval tasks (e.g. $39.8$ on R@1-i2t). Conversely, those have solid accuracy performance (e.g. $131.2$ on CIDEr) did achieve close performance on retrieval (e.g. $50.3$ on R@1-i2t).

3) The diverse performance does have some influence when the number of retrieval candidates increases. After RL training, captioning model preserves less effective candidates and the preserved candidates have a higher correlation level with images. Hence, the RL-trained model gets a higher score on R@1 but lower on R@5 and R@10. Our RMR baseline will supply this gap in Section~\ref{rmr}.

In short, if we want to generate human-like captions, we have to ensure the accuracy scores first. Otherwise, the generated captions can be semantic incorrect.

\section{Variational Transformer}

\subsection{Variational Auto Encoder}
\label{vae}

Typically, VAE theory was established on an assumption that the raw data points $x$ cluster around a low-dimensional manifold parameterized by embeddings $z$~\cite{Wang2017Diverse}. Thus, we may rebuild $x$ from $z$ as long as we know the true distribution of $z$. The right side of the Equation (\ref{eq1}) shows the Evidence Lower Bound (ELBO) on the log-likelihood of $x$ in the vanilla VAE~\cite{Kingma2014Auto-Encoding}. In the ideal case, we hope the distance between $p$ and $q$ to be minimized to $0$. Hence, to maximize the objective likelihood of $x$ , we only need to minimize the negative ELBO. However, the true distribution of $z$ is rather difficult to discover with limited data sources. A convensional solution is to assume that the latent variable $z$ behaves according to a given distribution, such as the standard normal distribution~\cite{Kingma2014Auto-Encoding} or the gaussian mixture distribution~\cite{Wang2017Diverse}.

\begin{equation}
\begin{aligned}
\log p(x)& - {D_{{\rm{KL}}}}\left[ {q\left( {z{\rm{|}}x} \right)\parallel p\left( {z{\rm{|}}x} \right)} \right]=\\
&{{\mathbb{E}}_{q\left( {z|x} \right)}}\left[ {\log p\left( {x{\rm{|}}z} \right)} \right] - {D_{{\rm{KL}}}}\left[ {q\left( {z{\rm{|}}x} \right)\parallel p\left( z \right)} \right].
\end{aligned}
\label{eq1}
\end{equation}

Reviewing the ELBO in Equation (\ref{eq1}), we found two specific optimization targets: the reconstructed log-likelihood of data point $x$ and the KL divergence between the posterior $q\left( {z{\rm{|}}x} \right)$ and the prior $p\left( x \right)$. In common AE models, we only employ the log-likelihood as the reconstruction loss, while in VAEs, the KL divergence guides an extra variational route based on the normal AE structure. In our model, we utilize this route to introduce the “Invisible Information Prior”.

\subsection{Overall Framework}
\label{vat}

In our VaT model, several Attention Layers and Samplers are employed to establish the deterministic and stochastic connections between input series and output probabilities. In Figure~\ref{fig1}, we have two different Attention Layers. For Self-Attention Layer, we retain the same structure of the original Transformer Encoder. For Multi-Head Attention Layer, we only make minor adjustments based on the original Transformer Multi-Head Attention module. In specific, we selectively compose the query searching, the residual structure and the feed forward module in the same layer for different parts, as shown in the right part of Figure~\ref{fig1}. In the following, we will introduce how we design the variational route and utilize the invisible language information to navigate our VaT model to manage the “trade-off” conflict.

\paragraph{Invisible Information Prior.}
The common language generation process employs a word-by-word pattern. For each timestep, the pdf of the current word $x_t$ is based on the generated sentence fragment $x_{<t}$. As shown in Equation (~\ref{eq1}), each word can only see the previous sentence fragment, which is incomplete. This partial visible problem is essential and hard to find a solution under the common AE architechture, due to the lacking of ground truths at inference time. Therefore, we consider using VAE to make the information loss recuperated.

As introduced in Section~\ref{vae}, the normal variational process has an extra prior route during training period. This prior route gives the potential for VAE to introduce the invisible information when using the word-by-word generating pattern. To be specific, for each position of the latent variable $z$, we regard the full target sentence $x$ as the prior information, meanwhile, regard the masked sentence $x_{<t}$ and the image $I$ as the posterior information to fit the generating pattern at inference time. Then we can reformulate the KL divergence of each position $t$ into ${D^t_{\rm{KL}}}\left[ {q\left( {z{\rm{|}}{x_{<t},I}} \right)\parallel p\left( {\tilde z}{\rm{|}}{x} \right)} \right]$. Under this alternation of target function, we will obtain a new variational route in our model, as shown in Figure~\ref{fig1}. At training time, we use the prior $\tilde z$ to decode the output probabilities for each word, while, at inference time, use $z$ instead. To navigate the posterior encoder to learn the invisible language information from prior $\tilde z$, we follow the VAE theory to minimize the KL divergence between pdfs of $z$ and $\tilde z$. Given an image $I$, we can draw the training target, i.e.,
\begin{equation}
\begin{aligned}
L {\rm =}&-{{\mathbb{E}}_{{q_\phi }\left( {z|{\tilde x},I} \right)}}\left[ {\log {p_\theta }\left( {x{\rm{|}}z,I} \right)} \right]\\
&+{D_{\rm{KL}}}\left[ {{q_\phi }\left( {z{\rm{|}}{\tilde x},I} \right)\parallel {p_\varphi }\left( {\tilde z{\rm{|}}x} \right)} \right],
\end{aligned}
\label{eq2}
\end{equation}
where $\theta$, $\phi$ and $\varphi$ denote the parameters in different modules, and,
\begin{equation}
\begin{aligned}
&{D_{\rm{KL}}}\left[ {{q_\phi }\left( {z{\rm{|}}{\tilde x},I} \right)\parallel {p_\varphi }\left( {\tilde z{\rm{|}}x} \right)} \right]\\
{\rm{ = }}&{\frac{1}{T}}\sum\limits_{t=1}^{T} {D^t_{\rm{KL}}}\left[ {{q_\phi}\left( {z{\rm{|}}{x_{<t},I}} \right)\parallel {p_\varphi}\left( {\tilde z}{\rm{|}}{x} \right)} \right].
\end{aligned}
\label{eq3}
\end{equation}
$T$ is the sequence length, and image $I$ here in the posterior ${q_\phi }\left( {z{\rm{|}}{\tilde x},I} \right)$ functions as a supplement for the posterior encoder to fill up the information loss caused by the fragmentary sentences.

\subsection{Auto-selectable Gaussian Mixture Model}
\label{auto_gmm}

\begin{figure}[]
    \centering
    \includegraphics[width=0.46\textwidth]{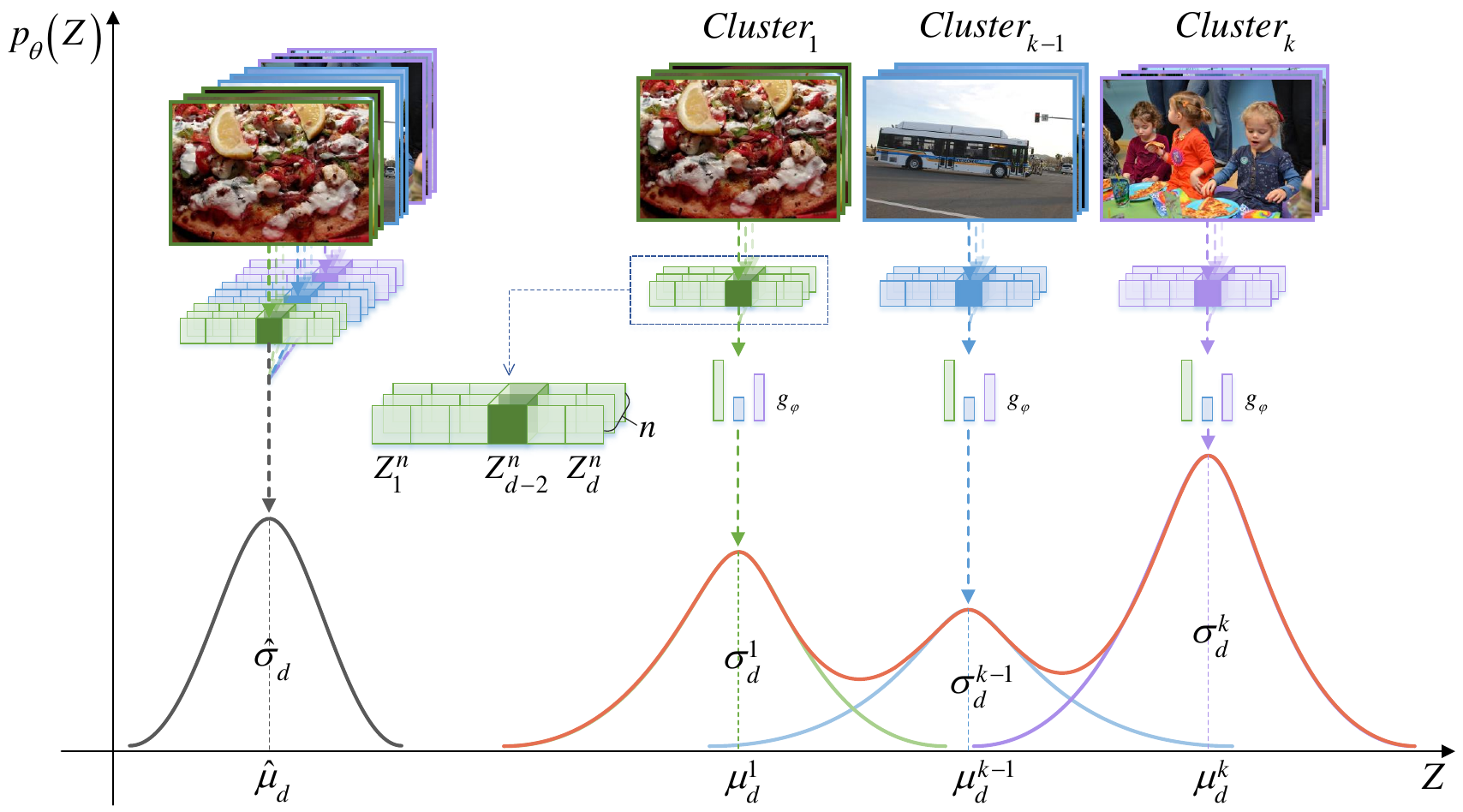}
    \caption{Comparison between the single gaussian prior and our auto-selectable GMM.}
    \label{fig2}
\end{figure}

In the classic VAE theory, we use the standard normal distribution as the hypothetical prior. However, as shown in Figure~\ref{fig2}, using single gaussian is trying to embed the information of the entire set of images into one tuple of parameter $(\mu, \sigma)$ for each dimension of $z$. This will, intuitively and practically, reserve much noisy information due to the indiscriminative embedding of images, meanwhile, raise a mismatch between the data distribution and the hypothesis prior. To overcome this problem, \cite{Wang2017Diverse} proposed a GMM-based CVAE model, in which the kernel of GMM was manually selected according to the object detection results of each image. This pattern has two issues. First, the capacity of GMM depends on the efficiency of the pre-trained detection model. Second, the same object in different scenes will share the same kernel in GMM, thus, have the same mean and variance. In another words, these objects with distinct semantic information will share the same latent representation. 

To solve this problem, we designed a novel GMM selection principal that can automatically match the object and its latent representation in different scenes using a simple learnable parameter $g_{\varphi}$. In specific, we choose the kernel for each dimension of $z$ according to the prior kernel probability $\omega{\rm =}{\rm softmax}(g_{\varphi})$. Technically, each dimension of $z$ will get a chance to fit the corresponding part of information into each kernel without force of mixture.

Given the auto-selection principle in Figure~\ref{fig2}, we follow the upper bound in \cite{Hershey2007Approximating} and make some slight modifications to turn it trainable end-to-end. Firstly, we consider $p$ and $q$ to be GMMs that have the same number of components $K$. The marginal densities of $x \in \mathbb{R}^d$ under $p$ and $q$ can be expressed as Equation~\ref{eq4}.
\begin{equation}
\begin{aligned}
&p\left( x \right) = \sum\limits_{k = 1}^K {{\omega _k}{\cal N}\left( {x;{\mu _k};{\Sigma _k}} \right)},\\
&q\left( x \right) = \sum\limits_{k = 1}^K {{{\tilde \omega }_k}{\cal N}\left( {x;{{\tilde \mu }_k};{{\tilde \Sigma }_k}} \right)},
\end{aligned}
\label{eq4}
\end{equation}
where ${\omega _k}$ and ${\tilde \omega _k}$ are the prior probabilities of each component in $p$ and $q$. ${\cal N}\left( {x;\mu ;\Sigma } \right)$ is a gaussian in $x$ with mean $\mu $ and covariance $\Sigma $. Then under the chain role of relative entropy~\cite{Cover1999Elements}, we have the following upper bound.
\begin{equation}
\begin{aligned}
{D_{\rm KL}}\left( {p||q} \right) &\le {D_{\rm KL}}\left( {\omega ||\tilde \omega } \right) + \sum\limits_{k = 1}^K {{\omega _k}{D_{\rm KL}}\left( {{p_k}||{q_k}} \right)}\\
&{\rm =} \sum\limits_{k = 1}^K {{\omega _k}\log \frac{{{\omega _k}}}{{{{\tilde \omega }_k}}}}  + \sum\limits_{k = 1}^K {{\omega _k}{D_{\rm KL}}\left( {{p_k}||{q_k}} \right)}.
\end{aligned}
\label{eq5}
\end{equation}

This upper bound can be further minimized by searching for the optimized mapping relation between the components of $p$ and $q$, but the searching process is too expensive for the deep learning model. Consequently, we replace the KL-divergence in Equation~\ref{eq2} with this practicable upper bound and reform our training target as:
\begin{equation}
\begin{aligned}
{L_{{\rm{vat}}}} = &- {{\mathbb{E}}_{q_\phi }}\left[ {\log {p_\theta }\left( {x{\rm{|}}z,{I}} \right)} \right] \\
&+ \beta  * \sum\limits_{k = 1}^K {{\omega _k}\left( {\log \frac{\omega _k}{{\tilde \omega }_k}+ D_{\rm{KL}}\left[ q_{\phi _k}\parallel p_{\varphi _k} \right]} \right)},
\end{aligned}
\label{eq6}
\end{equation}
where $\beta $ is the coefficient to adjust the ability of the proposed disentanglement theory~\cite{Higgins2017beta-VAE}. Through this reformulation, we can transform the calculation of the KL-divergence between GMMs into the calculation between each component of GMMs. Especially when we let every components in the upper bound~\ref{eq5} have the same prior probabilities as $\frac{1}{K}$, it will be equivalent to the expectation of the KL-divergence between each component pair, like the implementation in \cite{Wang2017Diverse}.

\subsection{Range-Median Reward Baseline}
\label{rmr}

The policy gradient of SCST shows in Equation~\ref{eq7}, where $\hat x = \left( {{\hat x_1}, \ldots ,{\hat x_T}} \right)$, $\hat x_t$ is the word sampled from the model at sequence position $t$, and $b$ is the greedy search baseline. This form introduces a better gradient variance reduction compared with the general cross-entropy loss and can improve accuracy dramatically. To achieve a further improvement, \cite{Luo2020Better} replaced the greedy sampled baseline $b$ with the average score of the rest sampled candidates. For $n$th sample, ${b_{avg}({\hat x_n})} = \mathop {\mathbb{E}}_{j \ne n} r\left( {{{\hat x}_j}} \right)$.

As indicated by ~\cite{shi2021off-policy}, SCST encourages the samples with higher scores to be more likely sampled along with the training progressed, which inevitably causes the diversity performance reduction. Even for our VaT model, using SCST will take a toll on the diversiy performance. Therefore, we proposed a novel baseline using the Range Median of all samples to improve the diversity without sacrificing the accuracy when adopting SCST method.

\begin{equation}
\begin{aligned}
{\nabla _\theta } \approx  - \left( {r\left( {\hat x} \right) - b} \right){\nabla _\theta }\log {p_\theta }\left( {\hat x|z,I} \right),
\end{aligned}
\label{eq7}
\end{equation}

\begin{figure}[]
  \begin{center}
    \includegraphics[width=0.46\textwidth]{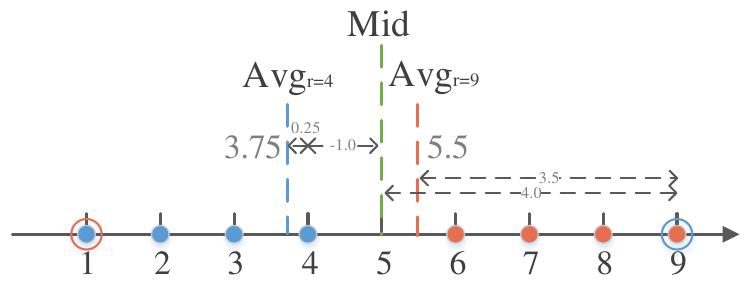}
    \caption{Intuitive analysis for different reinforce learning baselines.}
    \label{fig3}
  \end{center}
\end{figure}

We give an extreme case in Figure~\ref{fig3}, where the blue and orange circles indicate two groups of reward scores that have four close rewards and one outlier in each. We denote this two groups with $s_1=\{1, 2, 3, 4, 9\}$ and $s_2=\{1, 6, 7, 8, 9\}$. The green dashed line indicates our Range-Median Reward (RMR) baseline. The calculation formula shows in Equation~\ref{eq8}.

\begin{equation}
    \begin{aligned}
    b_{mid}\left(\hat x\right)=\left(\max \limits_{\hat x_n \in \hat X}\left(r \left(\hat x_n \right)\right)+\min \limits_{\hat x_n \in \hat X}\left(r \left(\hat x_n \right)\right)\right) /2
    \end{aligned}
    \label{eq8}
\end{equation}

In our formula, we consider the global information of all samples’ rewards. For example in group $s_1$, we have an extreme high score $9$. In the average baseline, the reward for the sample $4$ will be $0.25$, which encourages the sample with a low score. Meanwhile, in our median baseline, the reward for the sample $4$ decreases to $-1.0$, which properly punished the low-scored sample. 

Similarly, we will get a encouragement for sample $6$ in $s_2$ using our median baseline, while the average baseline will punish it. In this way, our median baseline reserves more positive samples with higher scores to improve the valid diversity performance instead keeping the false inferences for irrationally increasing the diversity metrics scores without considering the semantic accuracy. 

At last, we should note that, for the sample groups that have a balance distribution, our median baseline and the average baseline are less differentiating especially in the last stage of SCST training. The experimental results illustrate that our median baseline can get a fair accuracy performance compared with the average pattern, meanwhile, obtain a higher diversity performance.

\section{Evaluation}

\begin{table*}[]
\centering
\caption{Ablation study results on COCO Karpathy test split. Accuracy metrics: B-{\em N}, M, R, C, and S represent BLEU@{\em N}, METEOR, ROUGE-L, CIDEr, and SPICE. Diversity metrics: Uni., Div-{\em N}, mB-{\em N}, and All. represent Unique Sentence Ratio, n-gram diversity, mean BLEU-{\em N}, and AllSPICE. "$\downarrow$" denotes that the lower score is better. All results are reported in percentage(\%).}
\label{tabl: ablation}
\resizebox{\textwidth}{!}{%
\begin{tabular}{@{}cccccccccccccccc@{}}
\toprule
\multirow{4}{*}{\begin{tabular}[c]{@{}c@{}}Optimization\\ target\end{tabular}} &
\multirow{4}{*}{Models} &
  \multicolumn{6}{c}{\multirow{2}{*}{Accuracy}} &
  \multicolumn{6}{c}{\multirow{2}{*}{Diversity}} &
  \multirow{4}{*}{\begin{tabular}[c]{@{}c@{}}Accuracy \\ promoted\\ (average)\end{tabular}} &
  \multirow{4}{*}{\begin{tabular}[c]{@{}c@{}}Diversity\\ promoted\\ (average)\end{tabular}} \\
\multicolumn{2}{c}{} &
  \multicolumn{6}{c}{} &
  \multicolumn{6}{c}{} &
   &
   \\ \cmidrule(lr){3-14}
\multicolumn{2}{c}{} &
  \multirow{2}{*}{B-1} &
  \multirow{2}{*}{B-4} &
  \multirow{2}{*}{M} &
  \multirow{2}{*}{R} &
  \multirow{2}{*}{C} &
  \multirow{2}{*}{S} &
  \multirow{2}{*}{Uni.} &
  \multirow{2}{*}{All.} &
  \multirow{2}{*}{Div-1} &
  \multirow{2}{*}{Div-2} &
  \multirow{2}{*}{mB-4 $\downarrow$} &
  \multirow{2}{*}{S-C} &
   &
   \\
\multicolumn{2}{c}{} &
   &
   &
   &
   &
   &
   &
   &
   &
   &
   &
   &
   &
   &
   \\ \midrule
\multirow{3}{*}{CE} &
  Transformer(baseline) &
  75.6 &
  35.8 &
  27.9 &
  56.4 &
  114.0 &
  \textbf{21.1} &
  99.9 &
  22.3 &
  55.7 &
  75.6 &
  17.8 &
  89.8 &
  - &
  - \\
 &
  $\rm VaT_{GMM1}$ &
  75.9 &
  \textbf{35.9} &
  27.9 &
  56.5 &
  113.7 &
  20.8 &
  100.0 &
  22.1 &
  \textbf{59.2} &
  \textbf{79.6} &
  \textbf{12.4} &
  \textbf{92.0} &
  - &
  $\surd$ \\
 &
  $\rm VaT_{AGMM32}$ &
  \textbf{76.2} &
  35.8 &
  \textbf{27.9} &
  \textbf{56.5} &
  \textbf{114.4} &
  20.8 &
  \textbf{100.0} &
  \textbf{22.4} &
  58.6 &
  79.3 &
  12.6 &
  91.8 &
  $\surd$ &
  $\surd$ \\ \midrule
\multirow{3}{*}{CE+NSC} &
  Transformer(baseline) &
  80.8 &
  39.2 &
  29.0 &
  58.8 &
  130.1 &
  22.7 &
  54.6 &
  25.3 &
  22.6 &
  27.6 &
  89.0 &
  38.5 &
  - &
  - \\
 &
  $\rm VaT_{GMM1}$ &
  80.7 &
  39.4 &
  28.9 &
  58.8 &
  129.3 &
  22.6 &
  \textbf{73.9} &
  \textbf{26.8} &
  \textbf{26.8} &
  \textbf{34.6} &
  \textbf{79.9} &
  \textbf{46.8} &
  - &
  $\surd$ \\
 &
  $\rm VaT_{AGMM32}$ &
  \textbf{80.9} &
  \textbf{39.8} &
  \textbf{29.2} &
  \textbf{59.0} &
  \textbf{131.2} &
  \textbf{23.1} &
  70.1 &
  26.6 &
  25.4 &
  32.9 &
  82.3 &
  43.3 &
  $\surd$ &
  $\surd$ \\ \midrule
\multirow{3}{*}{CE+MSC} &
  Transformer(baseline) &
  80.7 &
  39.1 &
  29.0 &
  58.8 &
  130.1 &
  22.8 &
  63.1 &
  26.0 &
  24.0 &
  30.3 &
  85.6 &
  37.3 &
  - &
  - \\
 &
  $\rm VaT_{GMM1}$ &
  80.8 &
  39.5 &
  28.9 &
  58.8 &
  129.9 &
  22.7 &
  70.5 &
  26.3 &
  25.7 &
  33.4 &
  85.4 &
  44.1 &
  - &
  $\surd$ \\
 &
  $\rm VaT_{AGMM32}$ &
  \textbf{81.2} &
  \textbf{39.7} &
  \textbf{29.1} &
  \textbf{59.0} &
  \textbf{130.3} &
  \textbf{23.0} &
  \textbf{71.5} &
  \textbf{26.8} &
  \textbf{25.9} &
  \textbf{33.8} &
  \textbf{81.0} &
  \textbf{44.9} &
  $\surd$ &
  $\surd$ \\ \bottomrule
\end{tabular}%
}
\end{table*}

\subsection{Dataset and Evaluation Metrics}

\paragraph{Dataset}
We evaluate our model on the most popular benchmark MSCOCO~\cite{Lin2014Microsoft} in the image captioning area. For consensus comparison, we adopt the Karpathy’s split~\cite{Karpathy2015Deep}, which contains $113,287$ images for training, $5,000$ for validation and external $5,000$ for testing. Each image in the split is associated with at least five manual captions.

\paragraph{Accuracy metrics}
In our experiments, we follow the most popular AE methods to impose several metrics evaluate the quality of accuracy in contrast with the human captions, including BLEU~\cite{Papineni2002BLEU}, METEOR~\cite{Denkowski2014Meteor}, ROUGE~\cite{Lin2004Rouge:}, CIDEr~\cite{Vedantam2015Cider}, and SPICE~\cite{Anderson2016Spice}.

\paragraph{Diversity metrics}
For diversity evaluation, we adopt five benchmark diversity metrics in~\cite{Wang2017Diverse,Chen2019Variational,Luo2020Analysis,shi2021off-policy}. 1) {\em n-gram diversity} (Div-n): the ratio of distinct n-grams to the total number of words in the generated captions. Higher score of Div-n is better. 2) {\em mean Bleu-N} (mB-N): the mean value of the Bleu-N scores that are calculated between each caption in a set of K captions against the rest K-1 ones. It measures the inner similarities between the generated caption samples. Lower is better. 3) {\em Unique Sentence Ratio} (Uni.): the average ratio of distinct sentences in sampled sets. Higher is better. 4) {\em self-CIDEr} (S-C): singular vector decomposition (SVD) over autocorrelation matrices of the generated caption set using CIDEr as the kernel. Higher is better. 5) {\em AllSPICE} (All.): the F-score in a single scene graph for the generated caption set, that SPICE treats the same way with the reference caption sets. Higher is better with a balanced performance of accuracy and diversity.

\paragraph{Retrieval metrics}
For retrieval evaluation, we follow~\cite{Mao2015Deep, wang2020consensus} to adopt R@$K$ ($K=1, 5, 10$), which measures the fraction of queries for which the matched item is found among the top $K$ retrieved results, for both image-to-text and text-to-image evaluation. We also report the "mR" criterion that average all six recall rates of R@$K$. All reported retrieval results are experimented on Karpathy's 5K split, following~\cite{wang2020consensus}. 1K results can be found in Appendix.

\subsection{Implementation Details}

\paragraph{Image Feature Extractor}
To obtain the precise features corresponding to the ROIs under the guidance of Attention Mechanism, we follow the Updown method in~\cite{Anderson2018Bottom-Up} to use the pre-trained object features as image representation. For every image, we use a finetuned Faster R-CNN~\cite{Ren2015Faster} with Resnet-101~\cite{He2016Deep}, annotated and trained on the Visual Genome dataset~\cite{Anderson2018Bottom-Up,Krishna2017Visual}, to detect 10-100 regions (adaptive) and extract the corresponding features with 2048 dimensions. All the image features are pre-extracted as provided in~\cite{Anderson2018Bottom-Up,Luo2018Discriminability}.

\paragraph{Experiment Settings}
We set the batch size to $10$ in all our experiments for consensus. The number of layer was set to $4$, the inner-dimension $d$ as well as the dimension of the latent variable $z$ was set to $512$, and the dimension of the feedforward layer was $1024$. For every latent $z$ we set the number of GMM kernel to $32$. We trained our model with Adam optimization~\cite{Kingma2014Adam:} and the Reduce-LR-On-Plateau method for learning rate decay at every validation step. For initialization, we set the learning rate to $1 \times 10^{-4}$, the patience steps of decay to $3$, and the coefficient $\beta$ to $1.0$. The training process endured $30$ epochs including $15$ for the cross-entropy training and another $15$ for the self-critical training (using CIDEr optimization). Our project can be found on github~\footnote{https://github.com/kaelsunkiller/VaT\label{gitlink}}.

\subsection{Ablation Study}

To prove the effectiveness of our VaT framework and IIP module, we use the original Transformer in \cite{Vaswani2017Attention} as our baseline and follow its hyperparameter settings. To prove the effectiveness of our AGMM and RMR module, we use the single gaussian prior and the average reward RL baseline~\cite{Luo2020Better} as the contrast, separately. All models in our ablation experiments share the same hyperparameter settings and training strategy. In table~\ref{tabl: ablation}, we present both accuracy and diversity performance of different contrast models. CE, NSC and MSC indicate the cross entropy loss, the average self-critical optimization in \cite{Luo2020Better} and our RMR method, respectively. The subscript ${\rm GMM}N$ indicates that the model uses the GMM prior with $N$ kernels. We mark the best scores in bold and the second with the underline.

Under the same experiment conditions, our VaT framework using AGMM with $32$ kernels outperforms the Transformer baseline and simultaneously promotes the accuracy and diversity. Meanwhile, using the single gaussian prior can only promote diversity, as other diverse models do. Furthermore, experiments under NSC and MSC optimization illustrate that our RMR baseline maintains a better diversity performance without sacrificing the accuracy capacity (prevent the normal trade-off costs), especially when the trade-off gap is extremely exhibited by using the self-critical optimization.

\subsection{Evaluating the Accuracy Performance}

In Table~\ref{tabl: accuracy}, we introduce several state-of-the-art accurate methods, mainly including those based on the Transformer structure or similar attention oriented structures. Results from these methods with the superscript $*$ are reproduced under Luo's code framework~\footnote{https://github.com/ruotianluo/self-critical.pytorch\label{luogitlink}}, while others without $*$ are all quoted directly from the original papers. All models are trained with the self-critical optimization. Our model get a similar performance comparing with accurate models, especially when they are usually bad at generating diverse captions. In Section~\ref{section:diversiy}, We will show that our model performs outstandingly not only in accuracy evaluation but in diversity evaluation as well.
\begin{table}[]
\centering
\caption{Accuracy performance on COCO Karpathy test split, comparing with the state-of-the-art methods. All results are reported in percentage(\%).}
\label{tabl: accuracy}
\resizebox{0.47\textwidth}{!}{%
\begin{tabular}{@{}lcccccc@{}}
\toprule
Models                  & B-1           & B-4           & M             & R             & C              & S             \\ \midrule
$\rm Att2in^*$~\cite{Rennie2017Self-Critical}  & 78.4          & 35.7          & 27.3          & 56.9          & 119.5          & 20.7          \\
$\rm UpDown^*$~\cite{Anderson2018Bottom-Up} & 79.9          & 37.1          & 28.0          & 57.8          & 123.8          & 21.5          \\
$\rm AoA^*$~\cite{Huang2019Attention}     & 80.3          & 38.3          & 28.7          & 58.4          & 127.0          & 22.3          \\
SGAE~\cite{Yang2019Auto-encoding}            & 80.8          & 38.4          & 28.4          & 58.6          & 127.8          & 22.1          \\
$\rm Transformer^*$     & 80.8          & 39.2          & 29.0          & 58.8          & 130.1          & 22.7          \\
$\rm M2Transformer^*$~\cite{Cornia2020Meshed-Memory}   & 80.7          & 39.1          & 29.0          & 58.8          & 129.0          & 22.7          \\
B-SCST~\cite{Bujimalla2020B-SCST:}          & 80.8          & 39.0          & 29.2          & 59.0          & 131.0          & 22.9          \\
APN~\cite{yang2021apn}            & -             & 39.6          & 29.2          & \textbf{59.1} & \textbf{131.8} & 23.0          \\
MAC~\cite{ji2021mac}             & \textbf{81.5} & 39.5          & \textbf{29.3} & 58.9          & \underline{ 131.6}    & 22.8          \\
TRRL~\cite{nie2021trrl}            & \underline{ 81.4}    & 39.2          & 28.5          & 59.0          & 128.7          & 22.0          \\ \midrule
$\rm VaT_{msc}$ (ours)  & 81.2          & \underline{ 39.7}    & 29.1          & 59.0          & 130.3          & \underline{ 23.0}    \\
$\rm VaT_{nsc}$ (ours)  & 80.9          & \textbf{39.8} & \underline{ 29.2}    & \underline{ 59.0}    & 131.2          & \textbf{23.1} \\ \bottomrule
\end{tabular}%
}
\end{table}

\subsection{Evaluating the Diversity Performance Associated with Accuracy Metrics}
\label{section:diversiy}

\begin{table}[]
\centering
\caption{Diversity performance of different models correlated to accuracy metrics. All results are reported in percentage(\%).}
\label{tabl: diversity}
\resizebox{0.47\textwidth}{!}{%
\begin{tabular}{@{}lcccccc@{}}
\toprule
\multicolumn{1}{c}{\multirow{2}{*}{Models}} & \multicolumn{3}{c}{Accuracy}                   & \multicolumn{3}{c}{Diversity}                  \\ \cmidrule(l){2-7} 
\multicolumn{1}{c}{}          & B-4  & M    & C     & Uni. & mB-4 $\downarrow$ & S-C  \\ \midrule
\multicolumn{7}{l}{Accurate Models}                                                   \\ \midrule
$\rm Att2in^*$~\cite{Rennie2017Self-Critical}                        & 35.7 & 27.3 & 119.5 & 51.8 & 90.6              & 27.3 \\
$\rm UpDown^*$~\cite{Anderson2018Bottom-Up}                        & 37.1 & 28.0 & 123.8 & 56.5 & 89.0              & 31.9 \\
$\rm AoA^*$~\cite{Huang2019Attention}                           & 38.3 & 28.7 & 127.0 & 57.8 & 89.0              & 32.4 \\
$\rm Transformer^*$           & 39.2 & 29.0 & 130.1 & 54.6 & 89.0              & 38.5 \\
$\rm M2Transformer^*$~\cite{Cornia2020Meshed-Memory}             & 39.1 & 29.0 & 129.0 & 63.7 & 85.4              & 38.2 \\ 
$\rm VaT_{msc}$(ours)                       & \underline{ 39.7}    & \underline{ 29.1}    & \underline{ 130.3}    & \textbf{71.5}  & \textbf{81.0} & \textbf{44.9} \\
$\rm VaT_{nsc}$(ours)                             & \textbf{39.8} & \textbf{29.2} & \textbf{131.2} & \underline{ 70.1}     & \underline{ 82.3}    & \underline{ 43.3}    \\ \midrule
\multicolumn{7}{l}{Diverse models}                                                    \\ \midrule
GMM-CVAE~\cite{Wang2017Diverse}              & 18.9 & 21.7 & 78.5  & 90.9 & 45.6              & 70.7 \\
CapGAN~\cite{shetty2017speaking}                & 15.8 & 22.1 & 68.7  & 78.0 & 76.9              & 59.0 \\
Off-Policy~\cite{shi2021off-policy} ($\epsilon = 0.1$) & 26.5 & 24.5 & 89.9  & 92.0 & 54.0              & 69.3 \\
Off-Policy~\cite{shi2021off-policy} ($\epsilon = 0.9$) & 15.0 & 19.9 & 57.3  & 99.6 & 27.3              & 80.6 \\
$\rm VaT_{ce}$(ours)                              & \textbf{35.8} & \textbf{27.9} & \textbf{114.4} & \textbf{100.0} & \textbf{12.6} & \textbf{91.8} \\ \midrule
Human (leave-one-out)       & 19.5 & 24.1 & 87.8  & 100.0 & 19.5              & 88.6 \\ \bottomrule
\end{tabular}%
}
\end{table}

\begin{figure*}
    \centering
    \subfigure[]{
        \label{fig5.a}
        \includegraphics[height=0.35\textwidth]{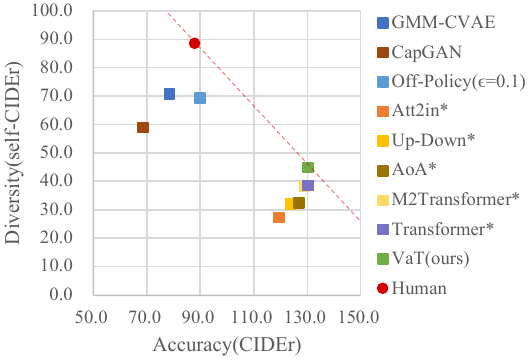}}
    \subfigure[]{
        \label{fig5.b}
        \includegraphics[height=0.35\textwidth]{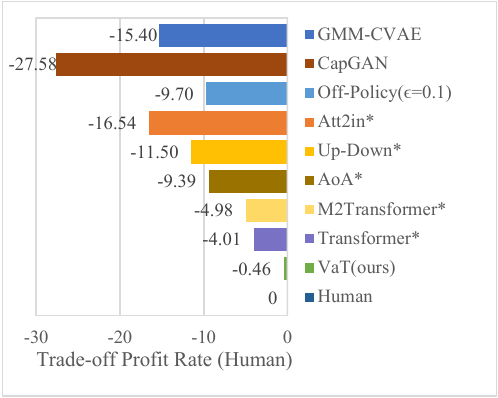}}
    \caption{Trade-off analysis of different works. (a) Performance of different works associated with both accuracy and diversity. The dashed line represent the zero TPR bound correlated to the human performance. (b) Relative Trade-off profit of each work. Our model achieves the closest result to human performance.All results are reported in percentage(\%).}
    \label{fig5}
\end{figure*}

Table~\ref{tabl: diversity} summarizes the diversity performance of different accurate and diverse models. We evaluate the diversity performance associated with the accuracy metrics to ensure that our model produces better diverse captions along with a solid accuracy performance. For consensus evaluation, all results of comparison methods are reported after the self-critical training. For multiple captions sampling, we employ the diverse sampling~\cite{Vijayakumar2016Diverse} with $\lambda=0.5$. 

First, we compare our model with AE models that have a better accuracy performance. Results suggest that our model outperforms others in both accuracy and diversity. Especially, we have achieved $44.9$ in self-CIDEr using the RMR baseline with $\mathbf{6.5}$ percent promotion compared with the best of AE models. 

Second, we compare our model with generative models that aim to promote the diversity. Note that we report the result of our model using CE optimization due to the inevitable diversity damage caused by the SCST training. Curiously, our model outperforms not only the other diverse models but also the leave-one-out results of human~\cite{Wang2019Describing}. Is that an evidence that our model generates better captions than human? Unfortunately, it is not. Actually, with CE optimization, most accurate models are able to provide better metric results than human's (leave-one-out). Naive Transformer gets $114.0$ in CIDEr and $89.8$ in self-CIDEr, for example. However, as we have indicated in Section~\ref{intro}, higher metric scores not consistently represent a better performance, a better diversity performance must establish on a solid accuracy performance to generate truly human-like captions. Then, how to measure the benefits of each model? In Section~\ref{section:eval-retrieval}, we evaluate both i2t and t2i retrieval for different models using a consensus pre-trained model. Also in Section~\ref{section:trade-off}, we propose a simple method to calculate model's capability of holding the diversity under a consensus criteria of accuracy.

\subsection{Evaluating the Semantic Correlation by Retrieval}
\label{section:eval-retrieval}

\begin{table}[]
\centering
\caption{Retrieval results on Karpathy's 5K test split. R@$K$($K=1, 5, 10$) is the rate of a correctly retrieved groundtruth given top $K$ candidates. mR is the mean value of all R@$K$. All results are reported in percentage(\%).}
\label{tabl: retieval}
\resizebox{0.47\textwidth}{!}{%
\begin{tabular}{@{}lccccccc@{}}
\toprule
\multicolumn{1}{c}{\multirow{2}{*}{Models}} & \multicolumn{3}{c}{Image-to-Text}                   & \multicolumn{3}{c}{Text-to-Image}           & \multirow{2}{*}{mR} \\ \cmidrule(lr){2-7}
\multicolumn{1}{c}{}    & R@1  & R@5  & R@10 & R@1  & R@5  & R@10 &      \\ \midrule
Human                   & 50.6 & 79.4 & 88.8 & 36.3 & 68   & 79.5 & 67.1 \\ \midrule
$\rm Att2in^*$~\cite{Rennie2017Self-Critical}  & 28.3 & 61.9 & 75.6 & 21.1 & 50.3 & 65.7 & 45.4 \\
$\rm UpDown^*$~\cite{Anderson2018Bottom-Up} & 36.2 & 57.4 & 68.6 & 26.5 & 58.5 & 73.1 & 53.4 \\
$\rm AoA^*$~\cite{Huang2019Attention}     & 42.9 & 63   & 72.9 & 31.6 & 64.2 & 77.4 & 58.7 \\
$\rm Transformer^*$     & 46.1 & 65.3 & 75.6 & 35.5 & 68.3 & 80.6 & 61.9 \\
$\rm M2Transformer^*$~\cite{Cornia2020Meshed-Memory}   & 47.3 & 68.9 & 78.5 & 34.4 & 66.9 & 79.4 & 62.6 \\ \midrule
$\rm VaT_{msc}$ (ours)                      & \underline{ 50.1}    & \textbf{74.1}       & \textbf{82.8} & \underline{ 35.9}    & \underline{ 69.2}  & \underline{ 81.8}    & \underline{ 65.7}          \\
$\rm VaT_{nsc}$ (ours)                      & \textbf{50.3} & \underline{ 73.1} & \underline{ 82.7}    & \textbf{36.7} & \textbf{70} & \textbf{82.2} & \textbf{65.8}       \\ \bottomrule
\end{tabular}%
}
\end{table}

Table~\ref{tabl: retieval} summarizes the retrieval results for both image-to-text and text-to-image evaluation. Our method achieves the best result and even surpass the human annotations on text-to-image retrieval. Note that this evaluation is based on a fixed pre-trained model (CVSE). We just replace the human annotations with captions genereated by different ML models. Every hyperparameters of CVSE are unmodified and the project can be found in the original paper~\cite{wang2020consensus}.

The interesting thing is that, by simply changing the annotations, captions generated by our model can improve the image retrieval performance up to $2.7$ percentage without finetuning the retrieval model. We suppose that the hypotheses from our model are more precise and easy for retrieval model to understand rather than human annotations. On the other hand, our model has better diversity performance, so that the promotion on R@10 is higher than that on R@1. Still and all, it does not mean that captions from our model outperform the human annotations. The utility of ML-based Retrieval is still circumscribed. We believe more effective methods will be proposed in the future.

\subsection{Analysing the Trade-off Gap with Human Performance}
\label{section:trade-off}

In order to generate diverse captions under the bondage of semantic accuracy, the self-critical training must be involved. Yet, the gap of accuracy between diverse models and accurate models becomes an intractable heterogeneity. To solve this problem, we need a consistent reference value, in which the margin of "trade-off" can be borrowed. From this conception, we introduce a simple measurement to calculate the "trade-off" using human performance as the reference.
\begin{equation}
    \begin{aligned}
    TPR\left( {\rm a}, {\rm b}\right) = \frac{1}{2} \left( \frac{Acc_{\rm a}-Acc_{\rm b}}{Acc_{\rm b}} + \frac{Div_{\rm a}-Div_{\rm b}}{Div_{\rm b}} \right)
    \end{aligned}
    \label{eq9}
\end{equation}
Equation~\ref{eq9} is the formulation of our compounded Trade-off Profit Rate (TPR), where $\rm a$ can be models we aim to assess, $\rm b$ is the baseline. In this function, we consider both increase or decrease of accuracy and diversity correlated to human performance. According to the trade-off phenomenon, with one item (accuracy or diversity) increases, another (diversity or accuracy) generally decreases. TPR calculates the compounded promotion rate within the trade-off margin. In our experiments, we use leave-one-out captions of human as $\rm b$, CIDEr as $Acc$ and self-CIDEr as $Div$.

In Figure~\ref{fig5.a}, we demonstrate the performance of different works associated with both accuracy and diversity performance. The red dashed line is the zero bound of $TPR_{human}$, where for every point on the line $TPR\left( {\rm point}, {\rm human}\right)=0$. Our model locates closest to this bound, which indicates that our model achieves almost the same rate of accuracy promotion as the diversity consume. In Figure~\ref{fig5.b} we report the specific TPR scores of each work. We are the closest one to the human standard with the solid accuracy performance.

\subsection{Qualitative Analysis}
\label{section:qualitative}
Due to the page limit, we put the qualitative analysis in Supplemental Material, Appendix A.

\section{Conclusion}
In this work, we propose a novel framework consist of different well-designed modules to ensure the diverse generation with the accurate semantic structure. First, we give the group of IIP and AGMM to guarantee the accuracy performance. Then, we give the RMR baseline to improve the quality of diverse generation based on a solid accuracy foundation. Extensive experiments suggest that our model achieves a simultaneous promotion in both accuracy and diversity. Furthermore, to evaluate the overall performance under the trade-off phenomenon, we propose a simple measurement to calculate the compounded trade-off rate. Also, we get the closest performance to the human annotations on semantic retrieval evaluation.

\bibliography{VaT}

\newpage

\title{Supplementary Material}

\section{Appendix A}
\subsection{Qualitative Analysis}
We sampled several images from Karpathy's test split. Results in Table~\ref{tab:qualitative} shows captions generated from different models. We use the Transformer trained by self-critical optimization ($130.1$ on CIDEr, $30.4$ on self-CIDEr and $-4.01$ on ${\rm TPR_human}$) as the accurate model and the VaT model with gmm number of 32 trained by cross entropy optimization ($114.4$ on CIDEr, $103.2$ on self-CIDEr and $16.95$ on $TPR_{\rm human}$) as the diverse model. Note that, compared with diverse models, accurate models trained by CE optimization are also capable to achieve both high accuracy and diversity scores associated with the human leave-one-out captions ($87.8$ on CIDEr and $88.6$ on self-CIDEr). However, as we have mentioned, the accuracy performance obtained through CE optimization can not ensure the semantic correctness. In Table~\ref{tab:qualitative}, it is intuitive to discover that the CE optimized model with only higher diversity performance is much easier to produce false inferences. Distinct captions generated for each image are annotated with different colors. 

\subsection{Trade-off Conversion Rate for RL Training}
\begin{figure}[htbp]
  \centering
    \includegraphics[width=0.4\textwidth]{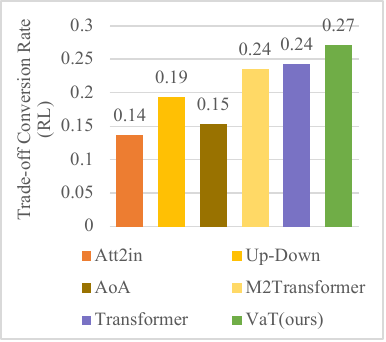}
    \caption{Trade-off conversion rate of different works.}
    \label{fig6}
\end{figure}

We also designed another variant of TPR, for evaluating the trade-off in RL training, i.e. Trade-off Conversion Rate (TCR). This perception is derived from the energy conversion efficiency (ECE): $\eta = P_{\rm out}/P_{\rm in}$. In our scenario, RL training can be considered as the machine that converse the input power (diversity) to the output power (accuracy). According to this concept, we built TCR to measure the trade-off efficiency of RL training for different methods. As shown in Figure~\ref{fig6}, our model achieves the best conversion performance.
\begin{equation}
\resizebox{0.4\textwidth}{!}{%
    $TCR_{\rm RL} = \left( \frac{\left|Div_{\rm CE}-Div_{\rm RL} \right|}{Div_{\rm RL}} \right) \bigg/ \left( \frac{\left|Acc_{\rm CE}-Acc_{\rm RL} \right|}{Acc_{\rm RL}} \right)$
}
\label{eq10}
\end{equation}

\subsection{Trade-off Analysis for CE-trained Models}
\label{sec:suppl.tprce}
\begin{figure*}
    \centering
    \subfigure[]{
        \label{fig.suppl.tprb}
        \includegraphics[height=0.35\textwidth]{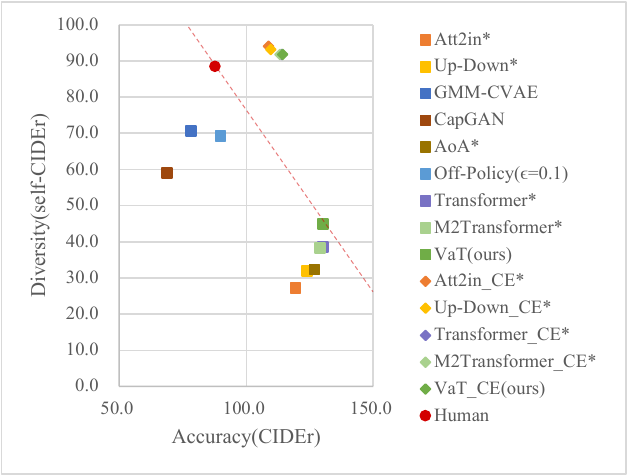}}
    \subfigure[]{
        \label{fig.suppl.tpra}
        \includegraphics[height=0.35\textwidth]{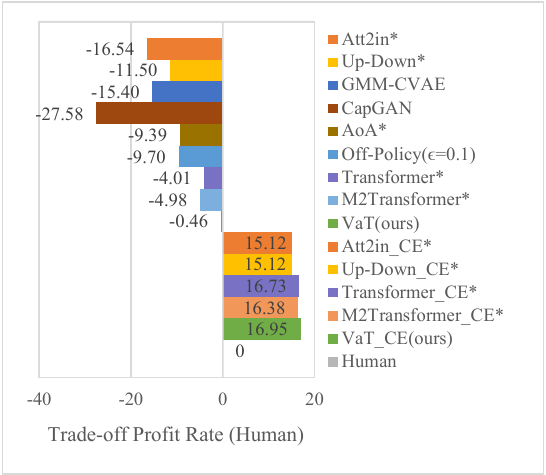}}
    \caption{Trade-off analysis including CE-trained models. (a) Performance of different works associated with both accuracy and diversity. The dashed line represent the zero TPR bound correlated to the human(leave-one-out) performance. (b) Relative Trade-off profit of each work. Our model achieves the closest result to human(leave-one-out) performance. All results are reported in percentage(\%).}
    \label{fig.suppl.tpr}
\end{figure*}

We reported the TRP score of each comparison method (trained by RL optimization) in the main content of this paper. Here, we also list the complete results including those trained by cross entropy loss, as shown in Figure~\ref{fig.suppl.tpr}.

It is confusing that if we only compare the accuracy and diversity performance with the human leave-one-out captions, we only need the cross-entropy training as they all exceeded the human zero bound by far. However, as we have proved empirically and experimentally, CE-trained models tended to generate more semantic errors to manifest inflating diversity scores and also did not performed well in retrieval tasks. Therefore, any simple measurement between accuracy and diversity may be failed when CE-trained models get involved. More complex measurement like retrieval should be considered under this circumstance. In Section \emph{Retrieval Analysis}, we will show that the result of retrieval evaluation is consistent with the intuitive experience for CE-trained models. Also, our TPR and TCR measurement still work for RL-trained models, as the result of retrieval confirms.

\subsection{Forward and Reverse KL-divergence}

It's known as the asymmetrical property of KL-divergence, that ${{D}_{\rm{KL}}}\left( P\parallel Q \right)\ne {{D}_{\rm{KL}}}\left( Q\parallel P \right)$, when approximating a true certain distribution $P\left( X \right)$ with an estimated distribution $Q\left( X \right)$ measured by the KL-divergence. Thus we have two kinds of formats --- forward (zero avoiding) and reverse (zero forcing) KL, in which we swap the position of $P$ and $Q$~\cite{fox_tutorial_2012}. In general VAEs, we use the reverse version, {\em i.e.}, ${{D}_{\rm{KL}}}\left( Q\parallel P \right)$. Different formats lead to different characteristics, in specific, the zero avoiding avoids $Q\left( x \right)=0$ whenever $P\left( x \right)>0$, and the zero forcing may force $Q\left( x \right)$ to be 0 even $P\left( x \right)>0$. This depends on the properties of KL-divergence that the probability weight of the difference between $P$ and $Q$ determines the sensibility of KL-divergence at the zero points. Especially for complex distributions $P$, the approximate distribution $Q$ will be extremely different when optimizing with these two forms. The forward KL tends to produce an "average" approximation, and the reverse KL tends to learn partial but more specific and accurate information. Note that this conclusion is under the case of a certain and aware true distribution $P$.

However, when we specify the $P$ and $Q$ as the single gaussian distribution, like in the vanilla VAE, the different optimization consequences between these two forms will be diminished. It's obvious that the unimodal distribution won't have the partial matching problem when measured by the KL-divergence. But in our model, the latent "instructor" and "leaner" distributions are not restricted as the single gaussian. It will cause the different characteristics we mentioned above if we use the rigorous KL-divergence between GMMs for optimization. Fortunately, we replace the KL-divergence between GMMs with an upper bound. This upper bound, in fact, aims to force each component-pair in GMMs to be fitted reciprocally by the KL-divergence. As each component is a single gaussian, we may get rid of the partial matching problem. The only thing we have to consider is the effect of gaussian prior weights. The KL-divergence between the prior weights should have different optimizations under forward and reverse mode, but in our experiments, we find it empirically also be unimodal for the distribution of the prior weights even without such restriction, which means it's unlike to have the partial matching problem for the prior weights either. 

\begin{table}[]
\centering
\caption{Additional results of Accuracy and Diversity metics on COCO Karpathy test split. oB4., oC., aC., oS. and aS. are short for Oracle BLEU-4 Oracle CIDEr, Average CIDEr, Oracle SPICE and Average SPICE, higher is better. All results are reported as percentage (\%). We generate $5$ sentences for each image, using naive sampling.}
\label{tab:suppladd}
\resizebox{\linewidth}{!}{%
\begin{tabular}{@{}lcccccc@{}}
\toprule
                     & \multicolumn{1}{l}{oB4.} & \multicolumn{1}{l}{aB4.} & oC.            & aC.            & oS.           & aS.           \\ \midrule
\multicolumn{7}{l}{Cross-entropy Optimization}                                   \\ \midrule
$\rm Att2in^*$~\cite{Rennie2017Self-Critical}        & 11.2 & 2.8  & 78.1  & 41.5  & 20.0 & 11.8 \\
$\rm UpDown^*$~\cite{Anderson2018Bottom-Up}        & 14.1 & 3.6  & 86.9  & 47.1  & 21.5 & 12.9 \\
$\rm AoA^*$~\cite{Huang2019Attention}           & 7.8  & 1.8  & 55.8  & 26.3  & 17.6 & 9.2  \\
$\rm Transformer^*$          & 20.2 & 5.8  & 100.8 & 56.5  & 23.5 & 14.4 \\
$\rm M2Transformer^*$~\cite{Cornia2020Meshed-Memory} & 20.6 & 6.0  & 101.8 & 56.6  & 23.4 & 14.3 \\
$\rm VaT$ (ours)     & \textbf{21.0}            & \textbf{6.1}             & \textbf{103.2} & \textbf{57.5}  & \textbf{24.0} & \textbf{14.9} \\ \midrule
\multicolumn{7}{l}{CIDEr Score Optimization}                             \\ \midrule
$\rm Att2in^*$~\cite{Rennie2017Self-Critical}        & 29.8 & 23.3 & 132.4 & 117.9 & 23.2 & 20.5 \\
$\rm UpDown^*$~\cite{Anderson2018Bottom-Up}        & 32.6 & 24.4 & 138.6 & 120.2 & 24.4 & 21.2 \\
$\rm AoA^*$~\cite{Huang2019Attention}           & 33.9 & 25.6 & 142.9 & 124.4 & 25.3 & 22.1 \\
$\rm Transformer^*$          & 34.4 & 26.2 & 144.7 & 126.8 & 25.6 & 22.4 \\
$\rm M2Transformer^*$~\cite{Cornia2020Meshed-Memory} & 36.0 & 26.0 & 145.3 & 123.0 & 26.1 & 22.1 \\
$\rm VaT_{msc}$ (ours) & 38.1                     & 25.8                     & \textbf{151.3} & 124.0          & \textbf{27.2} & 22.3          \\
$\rm VaT_{nsc}$ (ours) & \textbf{38.1}            & \textbf{26.2}            & 150.7          & \textbf{125.1} & 27.0          & \textbf{22.5} \\ \bottomrule
\end{tabular}%
}
\end{table}

Under such analysis and experimental experiences, we also tried a forward version for our model. It seems the forward version gives better performance on metrics in both accuracy and diversity evaluation. But the promotion is delicate. We believe when two distributions are both trainable and tend to adjust to each other, the optimizations of the forward and the reverse form will not vary too much. Even though, in our model, the forward form is more rational as the true distribution is now becoming a "learner" instead of an "instructor" to be approximated without self-adjustment. In consequence, the forward KL-divergence is worth to explore in variational inference.

\subsection{Additional Experimental Results}

Table~\ref{tab:suppladd} showcases the results of additional metrics. The oracle and average scores indicate the upper bound and average performance of generating accurate captions. Our model outperforms the others on the most metrics under the optimization of the cross-entropy loss and the CIDEr-based self-critical loss.

\begin{table}[]
\centering
\caption{Additional results of Accuracy and Diversity metics on COCO Karpathy test split. oC., aC., oS., aS., and Self. are short for Oracle CIDEr, Average CIDEr, Oracle SPICE, Average SPICE and Self-CIDEr, higher is better. All results are reported as percentage (\%). We generate $5$ sentences for each image, using naive sampling.}
\label{tab:supplallspice}
\resizebox{\linewidth}{!}{%
\begin{tabular}{lcccccc}
\hline
                                              & \multicolumn{1}{l}{All.} & \multicolumn{1}{l}{Obj.} & Rel.         & Attr. & Size         & Color         \\ \hline
\multicolumn{7}{l}{Cross-entropy Optimization}                                                                                                                    \\ \hline
$\rm Att2in^*$~\cite{Rennie2017Self-Critical} & 18.0                     & 36.0                     & 3.0          & 9.5   & 9.7          & 11.1          \\
$\rm UpDown^*$~\cite{Anderson2018Bottom-Up}   & 19.5                     & 38.2                     & 3.8          & 11.0  & 11.5         & 14.3          \\
$\rm AoA^*$~\cite{Huang2019Attention}         & 13.6                     & 29.1                     & 2.1          & 5.7   & 8.1          & 10.2          \\
$\rm Transformer^*$                           & 21.3                     & 39.5                     & 5.2          & 13.5  & 13.2         & \textbf{20.8} \\
$\rm M2Transformer^*$~\cite{Cornia2020Meshed-Memory} & 21.4          & 39.9          & 5.4          & 13.6          & 12.9          & 17.8          \\
$\rm VaT$ (ours)                                     & \textbf{22.4} & \textbf{41.7} & \textbf{5.7} & \textbf{14.6} & \textbf{15.3} & 20.0          \\ \hline
\multicolumn{7}{l}{CIDEr Score Optimization}                                                                                                              \\ \hline
$\rm Att2in^*$~\cite{Rennie2017Self-Critical} & 22.9                     & 42.0                     & 6.9          & 10.5  & 3.5          & 9.2           \\
$\rm UpDown^*$~\cite{Anderson2018Bottom-Up}   & 24.1                     & 43.4                     & 7.9          & 12.3  & 4.8          & 13.4          \\
$\rm AoA^*$~\cite{Huang2019Attention}         & 25.0                     & 44.4                     & 8.3          & 13.6  & 4.3          & 15.6          \\
$\rm Transformer^*$                           & 25.4                     & 44.8                     & 8.4          & 14.1  & 5.0          & 18.7          \\
$\rm M2Transformer^*$~\cite{Cornia2020Meshed-Memory} & 25.6          & 44.8          & 8.7          & 15.0          & 5.5           & 20.8          \\
$\rm VaT_{msc}$ (ours)                                 & \textbf{26.8} & \textbf{46.3} & 9.3          & \textbf{16.6} & 7.7           & \textbf{22.0} \\
$\rm VaT_{nsc}$ (ours)                          & 26.6                     & 46.2                     & \textbf{9.5} & 16.3  & \textbf{8.5} & 21.0          \\ \hline
\end{tabular}%
}
\end{table}

We report a breakdown of ALLSPICE over various subcategories in Table~\ref{tab:supplallspice}. Results with the superscript $*$ are reproduced by us. ALLSPICE is the F-score in a single scene graph for the generated caption set, that SPICE treats the same way with the reference caption sets, higher is better with a balanced performance of accuracy and diversity. Our model achieves the best performance under both CE and RL training.

\subsection{Retrieval Analysis}
\label{sec:suppl.retrieval}
\begin{table}[]
\centering
\caption{Retrieval results on Karpathy's 5K test split. We tested different models on both image-to-text and text-to-image tasks.}
\label{tab:suppl.retrieval5k}
\resizebox{\linewidth}{!}{%
\begin{tabular}{@{}lccccccc@{}}
\toprule
\multicolumn{1}{c}{\multirow{2}{*}{Models}} & \multicolumn{3}{c}{Image to text}             & \multicolumn{3}{c}{Text to Image}             & \multirow{2}{*}{mR} \\ \cmidrule(lr){2-7}
\multicolumn{1}{c}{}                                 & R@1  & R@5           & R@10          & R@1  & R@5  & R@10 &      \\ \midrule
Human                                                & 50.6 & 79.4          & 88.8          & 36.3 & 68.0 & 79.5 & 67.1 \\ \midrule
\multicolumn{8}{l}{Cross-entropy Optimization}                                                                          \\ \midrule
$\rm Att2in^*$~\cite{Rennie2017Self-Critical}        & 23.5 & 54.3          & 68.1          & 15.4 & 38.4 & 52.0 & 41.9 \\
$\rm UpDown^*$~\cite{Anderson2018Bottom-Up}          & 29.8 & 61.9          & 75.6          & 19.0 & 44.5 & 58.1 & 48.2 \\
$\rm AoA^*$~\cite{Huang2019Attention}                & 27.8 & 58.4          & 71.5          & 15.7 & 37.5 & 48.8 & 43.3 \\
$\rm Transformer^*$                                  & 39.8 & 72.1          & 83.1          & 25.3 & 52.5 & 64.8 & 59.6 \\
$\rm M2Transformer^*$~\cite{Cornia2020Meshed-Memory} & 38.4 & 70.3          & 82.2          & 23.8 & 51.3 & 63.8 & 55.0 \\
$\rm VaT$ (ours)                            & \textbf{43.6} & \textbf{77.6} & \textbf{87.1} & \textbf{29.2} & \textbf{59.4} & \textbf{72.0} & \textbf{61.5}       \\ \midrule
\multicolumn{8}{l}{CIDEr Score Optimization}                                                                            \\ \midrule
$\rm Att2in^*$~\cite{Rennie2017Self-Critical}        & 28.3 & 61.9          & 75.6          & 21.1 & 50.3 & 65.7 & 45.4 \\
$\rm UpDown^*$~\cite{Anderson2018Bottom-Up}          & 36.2 & 57.4          & 68.6          & 26.5 & 58.5 & 73.1 & 53.4 \\
$\rm AoA^*$~\cite{Huang2019Attention}                & 42.9 & 63.0          & 72.9          & 31.6 & 64.2 & 77.4 & 58.7 \\
$\rm Transformer^*$                                  & 46.1 & 65.3          & 75.6          & 35.5 & 68.3 & 80.6 & 61.9 \\
$\rm M2Transformer^*$~\cite{Cornia2020Meshed-Memory} & 47.3 & 68.9          & 78.5          & 34.4 & 66.9 & 79.4 & 62.6 \\
$\rm VaT_{msc}$ (ours)                               & 50.1 & \textbf{74.1} & \textbf{82.8} & 35.9 & 69.2 & 81.8 & 65.7 \\
$\rm VaT_{nsc}$ (ours)                      & \textbf{50.3} & 73.1          & 82.7          & \textbf{36.7} & \textbf{70.0} & \textbf{82.2} & \textbf{65.8}       \\ \bottomrule
\end{tabular}%
}
\end{table}

We follow~\cite{wang2020consensus} to evaluate the retrieval tasks on Karpathy's 5K and 1K split. Results are reported in Figure~\ref{tab:suppl.retrieval5k} and Figure~\ref{tab:suppl.retrieval1k}, respectively.

There is no doubt that our model outperforms others across-the-board. The point we need to focus is that, the results of CE-trained models corroborate the subjective observation in Section \emph{Trade-off Analysis for CE-trained Models}, which indicates that judging a model with simple standard drawn from human leave-one-out captions may not be robust. In contrast, the retrieval results of RL-trained models supply the conclusion drawn by our TPR and TCR measurement. Through these comprehensive experiments, we confirm the good applicability of our TPR and TCR under the RL-based optimizations.

\begin{table}[]
\centering
\caption{Retrieval results on Karpathy's 1K test split.}
\label{tab:suppl.retrieval1k}
\resizebox{\linewidth}{!}{%
\begin{tabular}{@{}llllllll@{}}
\toprule
\multicolumn{1}{c}{\multirow{2}{*}{Models}} &
  \multicolumn{3}{c}{Image to text} &
  \multicolumn{3}{c}{Text to Image} &
  \multicolumn{1}{c}{\multirow{2}{*}{mR}} \\ \cmidrule(lr){2-7}
\multicolumn{1}{c}{} &
  \multicolumn{1}{c}{R@1} &
  \multicolumn{1}{c}{R@5} &
  \multicolumn{1}{c}{R@10} &
  \multicolumn{1}{c}{R@1} &
  \multicolumn{1}{c}{R@5} &
  \multicolumn{1}{c}{R@10} &
  \multicolumn{1}{c}{} \\ \midrule
Human                                                & 75.1          & 95.4          & 98.2          & 61.8          & 89.2          & 94.9          & 85.8          \\ \midrule
\multicolumn{8}{l}{Cross-entropy Optimization}                                                                                                                       \\ \midrule
$\rm Att2in^*$~\cite{Rennie2017Self-Critical}        & 48.9          & 81.9          & 92.5          & 34.3          & 67.4          & 79.2          & 67.4          \\
$\rm UpDown^*$~\cite{Anderson2018Bottom-Up}          & 58.8          & 86.6          & 93.5          & 39.4          & 72.2          & 83.5          & 72.3          \\
$\rm AoA^*$~\cite{Huang2019Attention}                & 53.9          & 85.3          & 92.2          & 32.6          & 62.8          & 73.6          & 66.7          \\
$\rm Transformer^*$                                  & 67.3          & 92.6          & 97.8          & 47.2          & 78.1          & 86.9          & 78.4          \\
$\rm M2Transformer^*$~\cite{Cornia2020Meshed-Memory} & 65.4          & 92.7          & 97.9          & 45.8          & 77.7          & 86.6          & 77.7          \\
$\rm VaT$ (ours)                                     & \textbf{69.6} & \textbf{95.7} & \textbf{98.5} & \textbf{52.9} & \textbf{84.5} & \textbf{92.2} & \textbf{82.3} \\ \midrule
\multicolumn{8}{l}{CIDEr Score Optimization}                                                                                                                         \\ \midrule
$\rm Att2in^*$~\cite{Rennie2017Self-Critical}        & 55.2          & 72.9          & 85.0          & 45.5          & 82.5          & 93.0          & 72.4          \\
$\rm UpDown^*$~\cite{Anderson2018Bottom-Up}          & 60.8          & 80.9          & 89.1          & 53.1          & 86.8          & 95.3          & 77.7          \\
$\rm AoA^*$~\cite{Huang2019Attention}                & 64.4          & 83.3          & 91.3          & 56.6          & 89.2          & 96.4          & 80.2          \\
$\rm Transformer^*$                                  & 70.6          & 86.2          & 93.1          & 61.4          & 92.0          & 98.0          & 83.6          \\
$\rm M2Transformer^*$~\cite{Cornia2020Meshed-Memory} & 71.1          & 87.9          & 93.6          & 60.9          & 90.9          & 96.7          & 83.5          \\
$\rm VaT_{msc}$ (ours)                               & 74.5          & \textbf{92.0} & 96.1          & 62.4          & 92.8          & 97.7          & 85.9          \\
$\rm VaT_{nsc}$ (ours)                               & \textbf{74.6} & 91.4          & \textbf{96.8} & \textbf{63.0} & \textbf{92.8} & \textbf{98.0} & \textbf{86.1} \\ \bottomrule
\end{tabular}%
}
\end{table}

\begin{table*}[]
\centering
\caption{Qualitative results sampled from Karpathy's test split. TRP denotes the proposed Trade-off Profit Rate (Human).}
\label{tab:qualitative}
\scriptsize
\resizebox{\textwidth}{!}{%
\begin{tabular}{p{0.28\textwidth}<{\centering}p{0.21\textwidth}<{\centering}p{0.21\textwidth}<{\centering}p{0.21\textwidth}<{\centering}p{0.21\textwidth}<{\centering}}
\toprule
\normalsize Image &
\normalsize  Human annotations &
\normalsize  VaT(ours, msc) &
\normalsize  Diverse model &
\normalsize  Accurate model \\ \midrule
\multicolumn{1}{c}{\begin{minipage}[b]{0.6\columnwidth} \centering \raisebox{-.5\height}{\includegraphics[width=\linewidth]{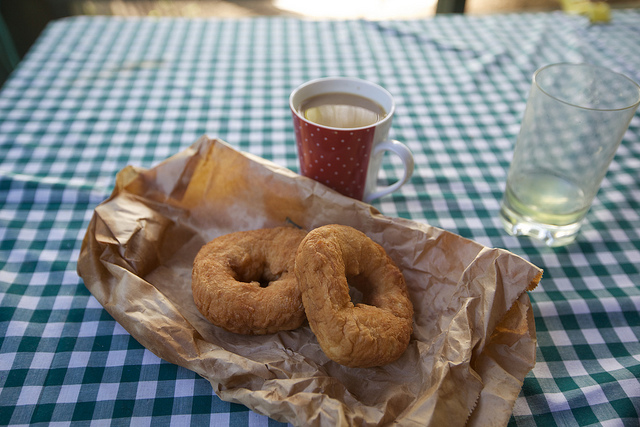}} \end{minipage}} &
  \begin{tabular}[l]{p{0.2\textwidth}}$\bullet$ \textcolor{sp1}{Two donuts sitting on top of a paper bag on a dinner table. }\\ $\bullet$ \textcolor{sp2}{Two donutes and a cup of coffee are placed on a table.}\\ $\bullet$ \textcolor{sp3}{Two doughnuts on top of a paper bag and a cup of coffee in a mug.}\\ $\bullet$ \textcolor{sp4}{Two doughnuts and a cup of coffee on a table.}\\ $\bullet$ \textcolor{sp5}{Two donuts are on the brown paper bag next to a cup of coffee. }\end{tabular} &
  \begin{tabular}[c]{p{0.2\textwidth}}$\bullet$ \textcolor{sp1}{two donuts and a cup of coffee on a table}\\ $\bullet$ \textcolor{sp2}{two donuts sitting on a table next to a cup of coffee}\\ $\bullet$ \textcolor{sp3}{three donuts sitting on a table with a cup of coffee}\\ $\bullet$ \textcolor{sp4}{three pastries on a table next to a cup of coffee}\\ $\bullet$ \textcolor{sp5}{two donuts on a piece of paper next to a cup of coffee}\end{tabular} &
  \begin{tabular}[c]{p{0.2\textwidth}}$\bullet$ \textcolor{sp1}{strawberry glazed doughnut sitting on a table next to a glass of drink}\\ $\bullet$ \textcolor{sp2}{two donuts and a cup of juice sit atop a table}\\ $\bullet$ \textcolor{sp3}{three doughnuts sit on a piece of napkins on straw in a bag next to a}\\ $\bullet$ \textcolor{sp4}{a plate containing sandwhich beside two glazed doughnuts and a cup of coffee are kept on}\\ $\bullet$ \textcolor{sp5}{a wooden table with two donuts and two glasses of beer}\end{tabular} &
  \begin{tabular}[c]{p{0.2\textwidth}}$\bullet$ \textcolor{sp1}{three donuts on a table with a cup of coffee}\\ $\bullet$ \textcolor{sp2}{two donuts on a table with a cup of coffee}\\ $\bullet$ \textcolor{sp3}{two donuts sitting on a table with two cups of coffee}\\ $\bullet$ \textcolor{sp4}{two donuts and two glasses of juice on a table}\\ $\bullet$ \textcolor{sp5}{three donuts sitting on a UNK next to a cup of coffee}\end{tabular}\\ \midrule
 &
   &
  \tiny CIDEr: 2.40, self-CIDEr: 0.66, TRP: 0.74 &
  \tiny CIDEr: 0.76, self-CIDEr: 0.94, TRP: -0.04 &
  \tiny CIDEr: 1.57, self-CIDEr: 0.73, TRP: 0.30 \\ \midrule
\multicolumn{1}{c}{\begin{minipage}[b]{0.6\columnwidth} \centering \raisebox{-.5\height}{\includegraphics[width=\linewidth]{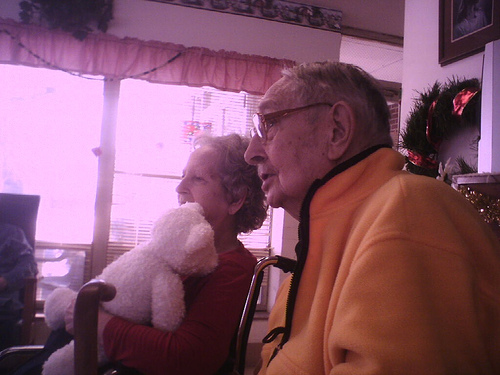}} \end{minipage}} &
  \begin{tabular}[c]{p{0.2\textwidth}}$\bullet$ \textcolor{sp1}{An elderly  man and woman look at something out of the camera's view. }\\ $\bullet$ \textcolor{sp2}{An older woman sitting in a wheel chair next to an older man.}\\ $\bullet$ \textcolor{sp3}{An old couple is sitting together holding a stuffed bear.}\\ $\bullet$ \textcolor{sp4}{The old man is sitting next to the old woman holding a teddy bear.}\\ $\bullet$ \textcolor{sp5}{ An elderly man and woman sitting down}\end{tabular} &
  \begin{tabular}[c]{p{0.2\textwidth}}$\bullet$ \textcolor{sp1}{a man and a woman holding a pink teddy bear}\\ $\bullet$ \textcolor{sp1}{a man and a woman holding a white teddy bear}\\ $\bullet$ \textcolor{sp2}{a man and a woman sitting with a pink teddy bear}\\ $\bullet$ \textcolor{sp3}{a man and a woman holding a pink teddy bear}\\ $\bullet$ \textcolor{sp4}{a man and a woman sitting in a chair with a pink teddy bear}\end{tabular} &
  \begin{tabular}[c]{p{0.2\textwidth}}$\bullet$ \textcolor{sp1}{two people and a girl embrace with a teddy bear}\\ $\bullet$ \textcolor{sp2}{a man holding a pink teddy bear next to a woman as they talk to her}\\ $\bullet$ \textcolor{sp3}{a man on his cell phone and a woman dressed up like a bunny}\\ $\bullet$ \textcolor{sp4}{a man and woman are sitting next to one of the young girl as they hold}\\ $\bullet$ \textcolor{sp5}{a couple of people that look like two ladies holding one eye with a pink head}\end{tabular} &
  \begin{tabular}[c]{p{0.2\textwidth}}$\bullet$ \textcolor{sp1}{a man and a woman sitting at a table with a teddy bear}\\ $\bullet$ \textcolor{sp1}{a man and a woman sitting at a table with a teddy bear}\\ $\bullet$ \textcolor{sp1}{a man and a woman sitting at a table with a teddy bear}\\ $\bullet$ \textcolor{sp1}{a man and a woman holding a teddy bear}\\ $\bullet$ \textcolor{sp2}{a man and a woman holding a teddy bear}\end{tabular}\\ \midrule
 &
   &
  \tiny CIDEr: 0.75, self-CIDEr: 0.47, TRP: -0.30 &
  \tiny CIDEr: 0.24, self-CIDEr: 0.95, TRP: -0.33 &
  \tiny CIDEr: 0.95, self-CIDEr: 0.27, TRP: -0.31 \\ \midrule
\multicolumn{1}{c}{\begin{minipage}[b]{0.6\columnwidth} \centering \raisebox{-.5\height}{\includegraphics[width=\linewidth]{samples/COCO_val2014_000000033052.jpg}} \end{minipage}} &
  \begin{tabular}[c]{p{0.2\textwidth}}$\bullet$ \textcolor{sp1}{a living room with a lot of chairs and a little bar in the corner }\\ $\bullet$ \textcolor{sp2}{A large living area with a fireplace and couches}\\ $\bullet$ \textcolor{sp3}{A living room has a large fireplace with decorative tiles.}\\ $\bullet$ \textcolor{sp4}{A living room with a fireplace and sofas are featured.}\\ $\bullet$ \textcolor{sp5}{A view of a living room area decorated with tiles and wood flooring.}\end{tabular} &
  \begin{tabular}[c]{p{0.2\textwidth}}$\bullet$ \textcolor{sp1}{a living room with a couch and a fireplace}\\ $\bullet$ \textcolor{sp1}{a living room with a fireplace and a table}\\ $\bullet$ \textcolor{sp1}{a living room with a living room with a fireplace}\\ $\bullet$ \textcolor{sp2}{a living room with a fireplace and a table}\\ $\bullet$ \textcolor{sp3}{a living room with a fireplace and a table}\end{tabular} &
  \begin{tabular}[c]{p{0.2\textwidth}}$\bullet$ \textcolor{sp1}{a bunch of furniture sits in a living room}\\ $\bullet$ \textcolor{sp2}{an organized living room has a fireplace and a fireplace}\\ $\bullet$ \textcolor{sp3}{a room with a fire place and other types of decorations on the wall}\\ $\bullet$ \textcolor{sp4}{a living area with a peanut butter fireplace and two couches}\\ $\bullet$ \textcolor{sp5}{an aerial view of a trestle chair table couch and sofa area}\end{tabular} &
  \begin{tabular}[c]{p{0.2\textwidth}}$\bullet$ \textcolor{sp1}{a living room with a fireplace and a couch}\\ $\bullet$ \textcolor{sp1}{a living room with a fireplace and a couch}\\ $\bullet$ \textcolor{sp1}{a living room with a fireplace and a couch}\\ $\bullet$ \textcolor{sp1}{a living room with a couch and a fireplace}\\ $\bullet$ \textcolor{sp2}{a living room with a couch and a fireplace}\end{tabular}\\ \midrule
 &
   &
  \tiny CIDEr: 2.24, self-CIDEr: 0.36, TRP: 0.48 &
  \tiny CIDEr: 0.67, self-CIDEr: 0.95, TRP: -0.08 &
  \tiny CIDEr: 2.19, self-CIDEr: 0.27, TRP: 0.40 \\ \midrule
\multicolumn{1}{c}{\begin{minipage}[b]{0.6\columnwidth} \centering \raisebox{-.5\height}{\includegraphics[width=\linewidth]{samples/COCO_val2014_000000553990.jpg}} \end{minipage}} &
  \begin{tabular}[c]{p{0.2\textwidth}}$\bullet$ \textcolor{sp1}{A young person ridding a horse jumps a gate in a competition.}\\ $\bullet$ \textcolor{sp2}{A man riding on a horse as it jumps over a pole. }\\ $\bullet$ \textcolor{sp3}{A woman is riding a horse as it jumps over a bar.}\\ $\bullet$ \textcolor{sp4}{there is a woman jockey riding a hose over the hurdle}\\ $\bullet$ \textcolor{sp5}{A woman riding a horse jumps over an obstacle.}\end{tabular} &
  \begin{tabular}[c]{p{0.2\textwidth}}$\bullet$ \textcolor{sp1}{a person jumping a horse over an obstacle}\\ $\bullet$ \textcolor{sp1}{a person riding a horse jumping over an obstacle}\\ $\bullet$ \textcolor{sp2}{a person riding a brown horse jumping over an obstacle}\\ $\bullet$ \textcolor{sp3}{a woman riding a horse jumping over an obstacle}\\ $\bullet$ \textcolor{sp4}{a woman riding a horse jumping over an obstacle}\end{tabular} &
  \begin{tabular}[c]{p{0.2\textwidth}}$\bullet$ \textcolor{sp1}{a horse jumping over an obstacle on a course}\\ $\bullet$ \textcolor{sp2}{a young girl riding a horse jumping over a barrier in a park}\\ $\bullet$ \textcolor{sp3}{a woman over a brown horse standing on top of a lush green field}\\ $\bullet$ \textcolor{sp4}{a horse rider jumping over a pink obstacle on grass with a safety net on its}\\ $\bullet$ \textcolor{sp5}{a girl jumping a show horse over a jump obstacle}\end{tabular} &
  \begin{tabular}[c]{p{0.2\textwidth}}$\bullet$ \textcolor{sp1}{a person riding a horse jumping over an obstacle}\\ $\bullet$ \textcolor{sp1}{a person riding a horse jumping over an obstacle}\\ $\bullet$ \textcolor{sp1}{a woman riding a horse jumping over an obstacle}\\ $\bullet$ \textcolor{sp2}{a woman riding a horse jumping over an obstacle}\\ $\bullet$ \textcolor{sp3}{a woman jumping a horse over a rail}\end{tabular}\\ \midrule
 &
   &
  \tiny CIDEr: 1.25, self-CIDEr: 0.44, TRP: -0.04 &
  \tiny CIDEr: 0.61, self-CIDEr: 0.87, TRP: -0.16 &
  \tiny CIDEr: 1.24, self-CIDEr: 0.35, TRP: -0.09 \\ \midrule
\multicolumn{1}{c}{\begin{minipage}[b]{0.6\columnwidth} \centering \raisebox{-.5\height}{\includegraphics[width=\linewidth]{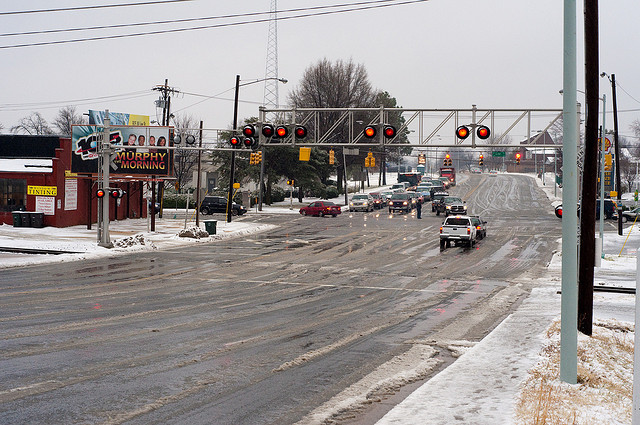}} \end{minipage}} &
  \begin{tabular}[c]{p{0.2\textwidth}}$\bullet$ \textcolor{sp1}{a number of cars stopped on a city street }\\ $\bullet$ \textcolor{sp2}{some lights are on over a road covered in snow}\\ $\bullet$ \textcolor{sp3}{a car drives down a snowy hil in the storm}\\ $\bullet$ \textcolor{sp4}{A street road with a bunch of water on it.}\\ $\bullet$ \textcolor{sp5}{The roads are a bit slushy by the train crossing.}\end{tabular} &
  \begin{tabular}[c]{p{0.2\textwidth}}$\bullet$ \textcolor{sp1}{a city street with traffic lights and a snow covered street}\\ $\bullet$ \textcolor{sp1}{a traffic light on a road in the snow}\\ $\bullet$ \textcolor{sp2}{a snow covered street with cars and a traffic light}\\ $\bullet$ \textcolor{sp3}{a traffic light on the side of a snow covered street}\\ $\bullet$ \textcolor{sp4}{a traffic light on the side of a snow covered street}\end{tabular} &
  \begin{tabular}[c]{p{0.2\textwidth}}$\bullet$ \textcolor{sp1}{a four way intersection of a slushy road area}\\ $\bullet$ \textcolor{sp2}{snow covered roads at an empty intersection with traffic lights on the road}\\ $\bullet$ \textcolor{sp3}{multiple stop lights covered in snow at an intersection with mountains nearby}\\ $\bullet$ \textcolor{sp4}{traffic is present as seen on a deserted road}\\ $\bullet$ \textcolor{sp5}{the traffic light is red on an empty street next to buildings}\end{tabular} &
  \begin{tabular}[c]{p{0.2\textwidth}}$\bullet$ \textcolor{sp1}{a city street with cars and traffic lights}\\ $\bullet$ \textcolor{sp1}{a city street with cars and traffic lights}\\ $\bullet$ \textcolor{sp2}{a city street with a traffic light and cars}\\ $\bullet$ \textcolor{sp3}{a snow covered street with a traffic light}\\ $\bullet$ \textcolor{sp4}{a transit limit with red traffic lights on a street}\end{tabular}\\ \midrule
 &
   &
  \tiny CIDEr: 0.33, self-CIDEr: 0.58, TRP: -0.48 &
  \tiny CIDEr: 0.24, self-CIDEr: 0.95, TRP: -0.33 &
  \tiny CIDEr: 0.37, self-CIDEr: 0.66, TRP: -0.42 \\ \midrule
\multicolumn{1}{c}{\begin{minipage}[b]{0.6\columnwidth} \centering \raisebox{-.5\height}{\includegraphics[width=\linewidth]{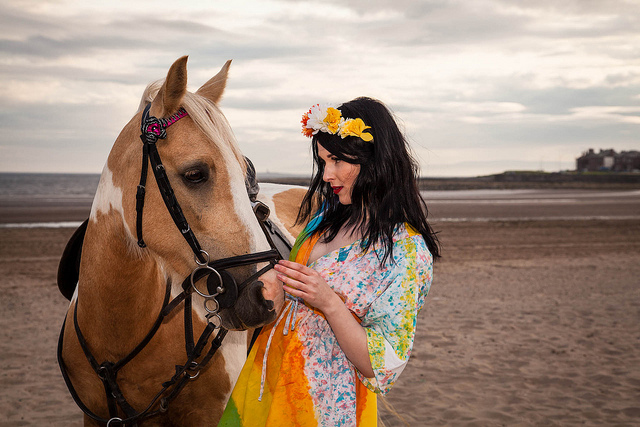}} \end{minipage}} &
  \begin{tabular}[c]{p{0.2\textwidth}}$\bullet$ \textcolor{sp1}{a woman with flowers in her hair staring at the horse next to her}\\ $\bullet$ \textcolor{sp2}{a person next to a horse on a beach}\\ $\bullet$ \textcolor{sp3}{The girl is holding the horse and looking at its face.  }\\ $\bullet$ \textcolor{sp4}{A girl with colorful clothing with a horse on a beach.}\\ $\bullet$ \textcolor{sp5}{A girl stands on the beach with a horse.}\end{tabular} &
  \begin{tabular}[c]{p{0.2\textwidth}}$\bullet$ \textcolor{sp1}{a woman standing next to a horse on the beach}\\ $\bullet$ \textcolor{sp1}{a woman standing next to a horse on the beach}\\ $\bullet$ \textcolor{sp1}{a woman standing next to a brown horse}\\ $\bullet$ \textcolor{sp2}{a woman standing next to a brown horse on the beach}\\ $\bullet$ \textcolor{sp3}{a woman standing next to a horse on the beach}\end{tabular} &
  \begin{tabular}[c]{p{0.2\textwidth}}$\bullet$ \textcolor{sp1}{a woman in a yellow dress and a horse}\\ $\bullet$ \textcolor{sp2}{a woman with bright hair stands next to a horse}\\ $\bullet$ \textcolor{sp3}{a woman and young girl in front of a saddled horse}\\ $\bullet$ \textcolor{sp4}{girl and mare hugging a horse on a dirt road}\\ $\bullet$ \textcolor{sp5}{woman with glasses with her horse along head and smiling for picture}\end{tabular} &
  \begin{tabular}[c]{p{0.2\textwidth}}$\bullet$ \textcolor{sp1}{a woman standing next to a horse on the beach}\\ $\bullet$ \textcolor{sp1}{a woman standing next to a horse on the beach}\\ $\bullet$ \textcolor{sp1}{a woman standing next to a horse on the beach}\\ $\bullet$ \textcolor{sp2}{a woman standing next to two horses on the beach}\\ $\bullet$ \textcolor{sp3}{a woman standing next to two horses on a beach}\end{tabular}\\ \midrule
 &
   &
  \tiny CIDEr: 1.83, self-CIDEr: 0.34, TRP: 0.23 &
  \tiny CIDEr: 0.86, self-CIDEr: 0.93, TRP: 0.02 &
  \tiny CIDEr: 1.59, self-CIDEr: 0.39, TRP: 0.12 \\ \bottomrule
\end{tabular}%
}
\end{table*}

\begin{table*}[]
\centering
\scriptsize
\resizebox{\textwidth}{!}{%
\begin{tabular}{p{0.28\textwidth}<{\centering}p{0.21\textwidth}<{\centering}p{0.21\textwidth}<{\centering}p{0.21\textwidth}<{\centering}p{0.21\textwidth}<{\centering}}
\toprule
\normalsize Image &
\normalsize  Human annotations &
\normalsize  VaT(ours, msc) &
\normalsize  Diverse model &
\normalsize  Accurate model \\ \midrule
\multicolumn{1}{c}{\begin{minipage}[b]{0.6\columnwidth} \centering \raisebox{-.5\height}{\includegraphics[width=\linewidth]{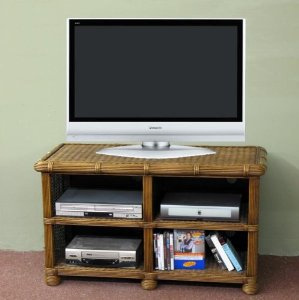}} \end{minipage}} &
  \begin{tabular}[c]{p{0.2\textwidth}}$\bullet$ \textcolor{sp1}{a tv on top of an entertainment center }\\ $\bullet$ \textcolor{sp2}{A plasma tv sits atop a small shelf.}\\ $\bullet$ \textcolor{sp3}{A television sitting on a stand with video devices.}\\ $\bullet$ \textcolor{sp4}{The television is on a wooden television stand}\\ $\bullet$ \textcolor{sp5}{A very simple entertainment center with a flatscreen TV}\end{tabular} &
  \begin{tabular}[c]{p{0.2\textwidth}}$\bullet$ \textcolor{sp1}{a flat screen tv sitting on top of a wooden book shelf}\\ $\bullet$ \textcolor{sp2}{a flat screen tv sitting on top of a wooden stand}\\ $\bullet$ \textcolor{sp3}{a television sitting on top of a wooden stand with books}\\ $\bullet$ \textcolor{sp4}{a television sitting on top of a book shelf}\\ $\bullet$ \textcolor{sp5}{a small entertainment center with a flat screen tv}\end{tabular} &
  \begin{tabular}[c]{p{0.2\textwidth}}$\bullet$ \textcolor{sp1}{a television is sitting on a stand with books}\\ $\bullet$ \textcolor{sp2}{a large flat screen tv sitting on top of a book shelf}\\ $\bullet$ \textcolor{sp3}{the entertainment table has a big screen monitor on it as well as a few books}\\ $\bullet$ \textcolor{sp4}{UNK in a neat living room in UNK fierce UNK of UNK UNK}\\ $\bullet$ \textcolor{sp5}{a tall entertainment stand displays many books and place sports movies UNK on the screen streetcar}\end{tabular} &
  \begin{tabular}[c]{p{0.2\textwidth}}$\bullet$ \textcolor{sp1}{a television on top of a book shelf}\\ $\bullet$ \textcolor{sp1}{a television on top of a book shelf}\\ $\bullet$ \textcolor{sp2}{a television on top of a book shelf filled with books}\\ $\bullet$ \textcolor{sp3}{a tv sitting on top of a wooden shelf}\\ $\bullet$ \textcolor{sp4}{a television is sitting on top of a entertainment center}\end{tabular}\\ \midrule
 &
   &
  \tiny CIDEr: 0.80, self-CIDEr: 0.75, TRP: -0.12 &
  \tiny CIDEr: 0.33, self-CIDEr: 0.95, TRP: -0.27 &
  \tiny CIDEr: 0.71, self-CIDEr: 0.57, TRP: -0.27 \\ \midrule
\multicolumn{1}{c}{\begin{minipage}[b]{0.6\columnwidth} \centering \raisebox{-.5\height}{\includegraphics[width=\linewidth]{samples/COCO_val2014_000000534121.jpg}} \end{minipage}} &
  \begin{tabular}[c]{p{0.2\textwidth}}$\bullet$ \textcolor{sp1}{A couch and a chair in a small room.}\\ $\bullet$ \textcolor{sp2}{a cozy room with leather furniture and a fireplace}\\ $\bullet$ \textcolor{sp3}{a fireplace a red couch two chairs and some lights}\\ $\bullet$ \textcolor{sp4}{A living room has plush furniture and a fireplace.}\\ $\bullet$ \textcolor{sp5}{This living room has leather furniture, a television, and a model ship on top of a fireplace. }\end{tabular} &
  \begin{tabular}[c]{p{0.2\textwidth}}$\bullet$ \textcolor{sp1}{a living room with a fireplace and a table}\\ $\bullet$ \textcolor{sp1}{a living room with a fireplace and a table}\\ $\bullet$ \textcolor{sp1}{a living room with a brick fireplace and a couch}\\ $\bullet$ \textcolor{sp2}{a living room with a fireplace and a table}\\ $\bullet$ \textcolor{sp3}{a living room with a brick fireplace and a fireplace}\end{tabular} &
  \begin{tabular}[c]{p{0.2\textwidth}}$\bullet$ \textcolor{sp1}{a view of a living room with seating}\\ $\bullet$ \textcolor{sp2}{this is an image of a living and room}\\ $\bullet$ \textcolor{sp3}{a living area with a wooden fire place and a mounted brick oven}\\ $\bullet$ \textcolor{sp4}{some light furniture sits in front of a fireplace and some plants}\\ $\bullet$ \textcolor{sp5}{a deep fireplace sitting inside of a room next to a fire place}\end{tabular} &
  \begin{tabular}[c]{p{0.2\textwidth}}$\bullet$ \textcolor{sp1}{a living room with a fireplace and a couch}\\ $\bullet$ \textcolor{sp1}{a living room with a fireplace and a couch}\\ $\bullet$ \textcolor{sp2}{a living room with a couch and a fireplace}\\ $\bullet$ \textcolor{sp3}{a living room with a fireplace and a chair}\\ $\bullet$ \textcolor{sp4}{a living room with a fireplace and a television}\end{tabular}\\ \midrule
 &
   &
  \tiny CIDEr: 1.23, self-CIDEr: 0.35, TRP: -0.10 &
  \tiny CIDEr: 0.43, self-CIDEr: 0.94, TRP: -0.22 &
  \tiny CIDEr: 1.59, self-CIDEr: 0.41, TRP: 0.14 \\ \midrule
\multicolumn{1}{c}{\begin{minipage}[b]{0.6\columnwidth} \centering \raisebox{-.5\height}{\includegraphics[width=\linewidth]{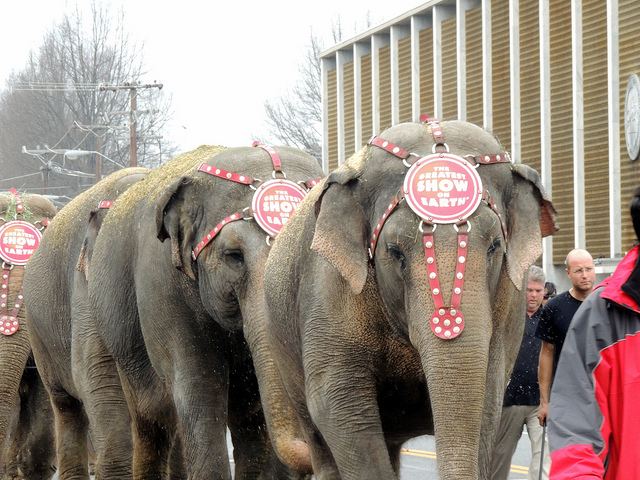}} \end{minipage}} &
  \begin{tabular}[c]{p{0.2\textwidth}}$\bullet$ \textcolor{sp1}{a couple of elephants are walking in a line}\\ $\bullet$ \textcolor{sp2}{Several elephants wearing decorations walking in a line.}\\ $\bullet$ \textcolor{sp3}{Two elephants that are standing up in the street.}\\ $\bullet$ \textcolor{sp4}{Several elephants walking in a line for a circus.}\\ $\bullet$ \textcolor{sp5}{Several elephants dressed for the circus are in line next to people. }\end{tabular} &
  \begin{tabular}[c]{p{0.2\textwidth}}$\bullet$ \textcolor{sp1}{a group of elephants with toothpicks on them}\\ $\bullet$ \textcolor{sp1}{a row of elephants with numbers on their ears}\\ $\bullet$ \textcolor{sp1}{a group of elephants standing next to each other}\\ $\bullet$ \textcolor{sp2}{a group of elephants standing next to each other}\\ $\bullet$ \textcolor{sp3}{a group of elephants standing next to each other}\end{tabular} &
  \begin{tabular}[c]{p{0.2\textwidth}}$\bullet$ \textcolor{sp1}{a herd of elephants walk in an exhibit while men interact}\\ $\bullet$ \textcolor{sp2}{a group of circus elephants standing with people looking on}\\ $\bullet$ \textcolor{sp3}{elephants with chains wearing number tags on their ears and sitting on a street}\\ $\bullet$ \textcolor{sp4}{elephants stand with dual faces on their faces a crowd looks on}\\ $\bullet$ \textcolor{sp5}{giant elephants are shown walking by walking down the street}\end{tabular} &
  \begin{tabular}[c]{p{0.2\textwidth}}$\bullet$ \textcolor{sp1}{a group of elephants standing next to a building}\\ $\bullet$ \textcolor{sp1}{a group of elephants standing next to a building}\\ $\bullet$ \textcolor{sp1}{a group of elephants standing next to a building}\\ $\bullet$ \textcolor{sp1}{a group of elephants standing next to a building}\\ $\bullet$ \textcolor{sp1}{a group of elephants standing next to a building}\end{tabular}\\ \midrule
 &
   &
  \tiny CIDEr: 0.87, self-CIDEr: 0.47, TRP: -0.24 &
  \tiny CIDEr: 0.64, self-CIDEr: 0.96, TRP: -0.10 &
  \tiny CIDEr: 1.15, self-CIDEr: 0.00, TRP: -0.35 \\ \midrule
\multicolumn{1}{c}{\begin{minipage}[b]{0.6\columnwidth} \centering \raisebox{-.5\height}{\includegraphics[width=0.8\linewidth]{samples/COCO_val2014_000000156292.jpg}} \end{minipage}} &
  \begin{tabular}[c]{p{0.2\textwidth}}$\bullet$ \textcolor{sp1}{A woman standing in front of an outdoor wall decorated with images and clocks.}\\ $\bullet$ \textcolor{sp2}{A person is looking at a display of clocks}\\ $\bullet$ \textcolor{sp3}{A man stands in front of a wall featuring a variety of dials, clocks, and pictures. }\\ $\bullet$ \textcolor{sp4}{A person standing looking at a  large statue with clocks built into it.}\\ $\bullet$ \textcolor{sp5}{a wall has many different clocks on display}\end{tabular} &
  \begin{tabular}[c]{p{0.2\textwidth}}$\bullet$ \textcolor{sp1}{a man standing in front of a wall with lots of clocks}\\ $\bullet$ \textcolor{sp2}{a man standing in front of a group of clocks on a wall}\\ $\bullet$ \textcolor{sp3}{a wall with a group of clocks on it}\\ $\bullet$ \textcolor{sp4}{a wall with a bunch of clocks on it}\\ $\bullet$ \textcolor{sp5}{a man standing in front of a bunch of clocks}\end{tabular} &
  \begin{tabular}[c]{p{0.2\textwidth}}$\bullet$ \textcolor{sp1}{a man standing in front of huge clocks on a wall with religious figures}\\ $\bullet$ \textcolor{sp2}{a giant clock on a wall in front of a building}\\ $\bullet$ \textcolor{sp3}{a group of clocks sit outside of a wall}\\ $\bullet$ \textcolor{sp4}{there are many clocks and clocks in this building}\\ $\bullet$ \textcolor{sp5}{five circular designed clocks mounted to the side of an old building}\end{tabular} &
  \begin{tabular}[c]{p{0.2\textwidth}}$\bullet$ \textcolor{sp1}{a group of people standing in front of a wall with clocks}\\ $\bullet$ \textcolor{sp1}{a group of people standing in front of a wall with clocks}\\ $\bullet$ \textcolor{sp1}{a group of clocks on the side of a building}\\ $\bullet$ \textcolor{sp2}{a group of clocks on the side of a building}\\ $\bullet$ \textcolor{sp3}{a bunch of clocks on a wall}\end{tabular}\\ \midrule
 &
   &
  \tiny CIDEr: 1.42, self-CIDEr: 0.61, TRP: 0.15 &
  \tiny CIDEr: 0.67, self-CIDEr: 0.91, TRP: -0.10 &
  \tiny CIDEr: 1.35, self-CIDEr: 0.52, TRP: 0.06 \\ \midrule
\multicolumn{1}{c}{\begin{minipage}[b]{0.6\columnwidth} \centering \raisebox{-.5\height}{\includegraphics[width=\linewidth]{samples/COCO_val2014_000000286981.jpg}} \end{minipage}} &
  \begin{tabular}[c]{p{0.2\textwidth}}$\bullet$ \textcolor{sp1}{A red hair woman holding an open box of pizza.}\\ $\bullet$ \textcolor{sp2}{A young woman holding a pizza in a box.}\\ $\bullet$ \textcolor{sp3}{a woman is holding a box of pizza}\\ $\bullet$ \textcolor{sp4}{A woman is posing with an open pizza box.}\\ $\bullet$ \textcolor{sp5}{A woman holds an open box of pizza. }\end{tabular} &
  \begin{tabular}[c]{p{0.2\textwidth}}$\bullet$ \textcolor{sp1}{a woman holding a box of pizza on a table}\\ $\bullet$ \textcolor{sp2}{a woman holding a box with a large pizza}\\ $\bullet$ \textcolor{sp3}{a woman holding a in front of a large pizza}\\ $\bullet$ \textcolor{sp4}{a woman holding a box of pizza sitting on a table}\\ $\bullet$ \textcolor{sp5}{a woman holding a large pizza in a box}\end{tabular} &
  \begin{tabular}[c]{p{0.2\textwidth}}$\bullet$ \textcolor{sp1}{a woman sits behind a large pizza made in a box}\\ $\bullet$ \textcolor{sp2}{a lady holding a pizza sitting next to a box of UNK on a table}\\ $\bullet$ \textcolor{sp3}{woman holding open two large pizzas with each other}\\ $\bullet$ \textcolor{sp4}{woman standing holding up a large pizza box}\\ $\bullet$ \textcolor{sp5}{this girl is enjoying her pizza at the table}\end{tabular} &
  \begin{tabular}[c]{p{0.2\textwidth}}$\bullet$ \textcolor{sp1}{a woman holding a large pizza in a table}\\ $\bullet$ \textcolor{sp2}{a woman holding a pizza in a table}\\ $\bullet$ \textcolor{sp3}{a woman sitting at a table with a large pizza}\\ $\bullet$ \textcolor{sp4}{a woman holding a box of pizza}\\ $\bullet$ \textcolor{sp5}{a woman sitting at a table with a box of pizza}\end{tabular}\\ \midrule
 &
   &
  \tiny CIDEr: 2.73, self-CIDEr: 0.63, TRP: 0.91 &
  \tiny CIDEr: 1.09, self-CIDEr: 0.91, TRP: 0.13 &
  \tiny CIDEr: 2.39, self-CIDEr: 0.69, TRP: 0.75 \\ \midrule
\multicolumn{1}{c}{\begin{minipage}[b]{0.6\columnwidth} \centering \raisebox{-.5\height}{\includegraphics[width=\linewidth]{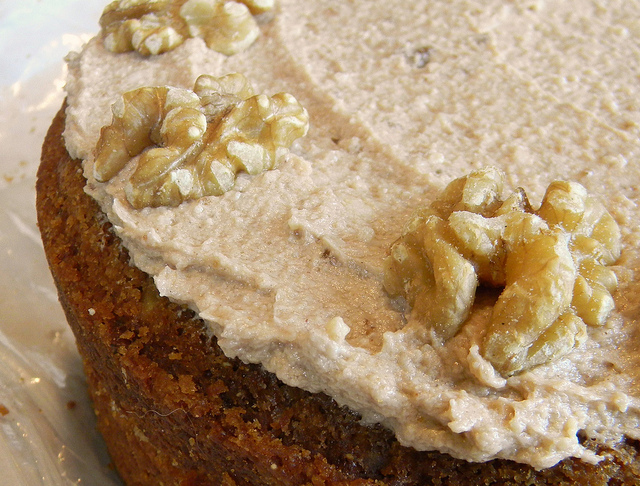}} \end{minipage}} &
  \begin{tabular}[c]{p{0.2\textwidth}}$\bullet$ \textcolor{sp1}{a brown cake with white icing and some walnut toppings}\\ $\bullet$ \textcolor{sp2}{Icing with nuts on top of a cake . }\\ $\bullet$ \textcolor{sp3}{Corner of a layer cake with frosting and walnuts.}\\ $\bullet$ \textcolor{sp4}{A frosted large cake on a table surface.}\\ $\bullet$ \textcolor{sp5}{Walnut icing on a baked cake - maybe carrot cake}\end{tabular} &
  \begin{tabular}[c]{p{0.2\textwidth}}$\bullet$ \textcolor{sp1}{a piece of cake with bananas on top of it}\\ $\bullet$ \textcolor{sp2}{a doughnut with cookie slices on top of it}\\ $\bullet$ \textcolor{sp3}{a piece of bread with bananas and nuts on top}\\ $\bullet$ \textcolor{sp4}{a piece of fruit with nuts on top of it}\\ $\bullet$ \textcolor{sp5}{cookies and bananas on a piece of bread on a plate}\end{tabular} &
  \begin{tabular}[c]{p{0.2\textwidth}}$\bullet$ \textcolor{sp1}{a muffin dessert desert with UNK and custard cream}\\ $\bullet$ \textcolor{sp2}{a stock container of powdered sugar banana slices and topped with thick sugar}\\ $\bullet$ \textcolor{sp3}{some type of bread covered in powdered sugar and peanut butter}\\ $\bullet$ \textcolor{sp4}{cinnamon roll desert on a plastic plate on a table}\\ $\bullet$ \textcolor{sp5}{cake with powdered sugar on top of it}\end{tabular} &
  \begin{tabular}[c]{p{0.2\textwidth}}$\bullet$ \textcolor{sp1}{a piece of bread with bananas on top of it}\\ $\bullet$ \textcolor{sp1}{a piece of bread with bananas on top of it}\\ $\bullet$ \textcolor{sp1}{a piece of bread with bananas on top of it}\\ $\bullet$ \textcolor{sp1}{a piece of bread with bananas on top of it}\\ $\bullet$ \textcolor{sp2}{a piece of bread with bananas on top}\end{tabular}\\ \midrule
 &
   &
  \tiny CIDEr: 0.39, self-CIDEr: 0.84, TRP: -0.30 &
  \tiny CIDEr: 0.10, self-CIDEr: 0.91, TRP: -0.43 &
  \tiny CIDEr: 0.05, self-CIDEr: 0.07, TRP: -0.93 \\ \bottomrule
\end{tabular}%
}
\end{table*}

\begin{table*}[]
\centering
\scriptsize
\resizebox{\textwidth}{!}{%
\begin{tabular}{p{0.28\textwidth}<{\centering}p{0.21\textwidth}<{\centering}p{0.21\textwidth}<{\centering}p{0.21\textwidth}<{\centering}p{0.21\textwidth}<{\centering}}
\toprule
\normalsize Image &
\normalsize  Human annotations &
\normalsize  VaT(ours, msc) &
\normalsize  Diverse model &
\normalsize  Accurate model \\ \midrule
\multicolumn{1}{c}{\begin{minipage}[b]{0.6\columnwidth} \centering \raisebox{-.5\height}{\includegraphics[width=\linewidth]{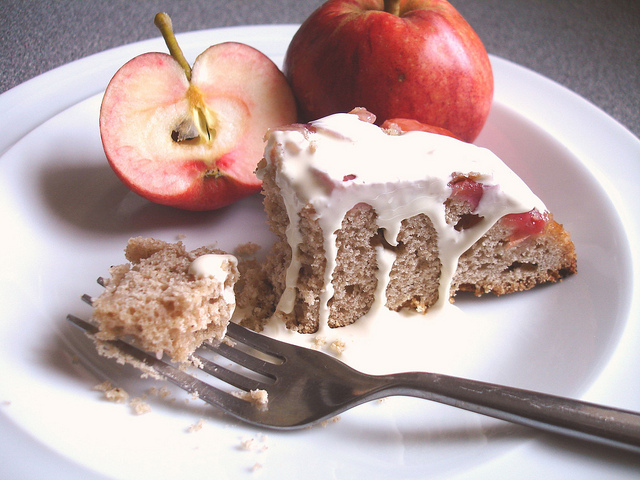}} \end{minipage}} &
  \begin{tabular}[c]{p{0.2\textwidth}}$\bullet$ \textcolor{sp1}{A desert with icing and a sliced apple beside it.}\\ $\bullet$ \textcolor{sp2}{A piece of cake with a fork and one and a half apples on the plate.}\\ $\bullet$ \textcolor{sp3}{An fresh apple and a half on a plate with a slice of apple bread dribbled with cream}\\ $\bullet$ \textcolor{sp4}{A dessert with cake and apples is on a plate with a fork.}\\ $\bullet$ \textcolor{sp5}{The cut apples give a hint about what the cake tastes like.}\end{tabular} &
  \begin{tabular}[c]{p{0.2\textwidth}}$\bullet$ \textcolor{sp1}{a piece of cake on a white plate with an apple}\\ $\bullet$ \textcolor{sp1}{a piece of cake on a plate with an apple}\\ $\bullet$ \textcolor{sp2}{a piece of cake on a white plate with an apple}\\ $\bullet$ \textcolor{sp3}{a white plate with two apples and a fork on it}\\ $\bullet$ \textcolor{sp4}{a white plate with a piece of cake and an apple}\end{tabular} &
  \begin{tabular}[c]{p{0.2\textwidth}}$\bullet$ \textcolor{sp1}{a UNK of a dessert nut and an apple artfully balanced on the plate}\\ $\bullet$ \textcolor{sp2}{a plate with an apple and two cherries on it}\\ $\bullet$ \textcolor{sp3}{a close up of a plate with food on it}\\ $\bullet$ \textcolor{sp4}{three pieces of square cake on a plate with a fork}\\ $\bullet$ \textcolor{sp5}{pieces of an apple sit on top of a white plate with a sprinkles hit and}\end{tabular} &
  \begin{tabular}[c]{p{0.2\textwidth}}$\bullet$ \textcolor{sp1}{a piece of cake on a plate with an apple}\\ $\bullet$ \textcolor{sp1}{a piece of cake on a plate with an apple}\\ $\bullet$ \textcolor{sp1}{a piece of cake on a plate with an apple}\\ $\bullet$ \textcolor{sp1}{a piece of cake on a plate with an apple}\\ $\bullet$ \textcolor{sp2}{a piece of cake on a white plate with an apple}\end{tabular}\\ \midrule
 &
   &
  \tiny CIDEr: 0.80, self-CIDEr: 0.51, TRP: -0.26 &
  \tiny CIDEr: 0.41, self-CIDEr: 0.93, TRP: -0.24 &
  \tiny CIDEr: 0.80, self-CIDEr: 0.13, TRP: -0.47 \\ \midrule
\multicolumn{1}{c}{\begin{minipage}[b]{0.6\columnwidth} \centering \raisebox{-.5\height}{\includegraphics[width=\linewidth]{samples/COCO_val2014_000000543300.jpg}} \end{minipage}} &
  \begin{tabular}[c]{p{0.2\textwidth}}$\bullet$ \textcolor{sp1}{A boat sits on the side of the dock.}\\ $\bullet$ \textcolor{sp2}{A large white boat in the open water.}\\ $\bullet$ \textcolor{sp3}{A white double decker boat n water next to buildings.}\\ $\bullet$ \textcolor{sp4}{A large cruise ship is traveling on the ocean. }\\ $\bullet$ \textcolor{sp5}{A Port River Dolphin Cruise ship sits in the water.}\end{tabular} &
  \begin{tabular}[c]{p{0.2\textwidth}}$\bullet$ \textcolor{sp1}{a boat in the water next to a building}\\ $\bullet$ \textcolor{sp2}{a cruise ship docked in front of a UNK}\\ $\bullet$ \textcolor{sp3}{a large boat docked in front of a building}\\ $\bullet$ \textcolor{sp4}{a boat in the water next to a dock}\\ $\bullet$ \textcolor{sp5}{a boat is parked in the water next to a building}\end{tabular} &
  \begin{tabular}[c]{p{0.2\textwidth}}$\bullet$ \textcolor{sp1}{a steamboat moving across the water near large buildings}\\ $\bullet$ \textcolor{sp2}{a large white boat is parked next to a large white house}\\ $\bullet$ \textcolor{sp3}{a cruise ship is in the harbor by some other large buildings}\\ $\bullet$ \textcolor{sp4}{white boat moored at dock with beach building in background}\\ $\bullet$ \textcolor{sp5}{it is hard to UNK in daytime on the water next to a town also port}\end{tabular} &
  \begin{tabular}[c]{p{0.2\textwidth}}$\bullet$ \textcolor{sp1}{a boat is sitting in the water next to a building}\\ $\bullet$ \textcolor{sp1}{a boat is in the water next to a building}\\ $\bullet$ \textcolor{sp2}{a boat is in the water next to a building}\\ $\bullet$ \textcolor{sp3}{a boat is in front of a building}\\ $\bullet$ \textcolor{sp4}{a boat is sitting in the water next to a that with a house}\end{tabular}\\ \midrule
 &
   &
  \tiny CIDEr: 0.88, self-CIDEr: 0.69, TRP: -0.11 &
  \tiny CIDEr: 0.48, self-CIDEr: 0.97, TRP: -0.18 &
  \tiny CIDEr: 0.82, self-CIDEr: 0.48, TRP: -0.26 \\ \midrule
\multicolumn{1}{c}{\begin{minipage}[b]{0.6\columnwidth} \centering \raisebox{-.5\height}{\includegraphics[width=0.6\linewidth]{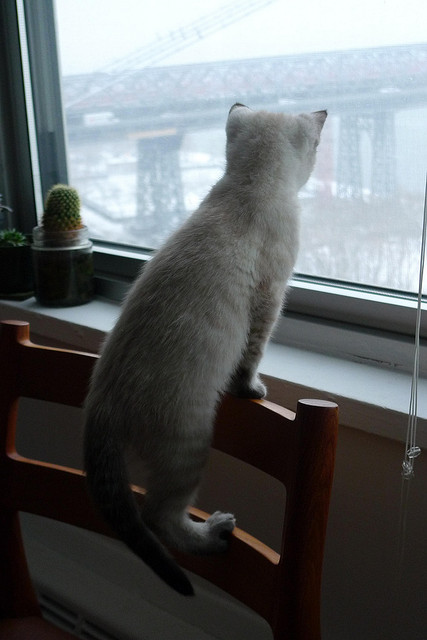}} \end{minipage}} &
  \begin{tabular}[c]{p{0.2\textwidth}}$\bullet$ \textcolor{sp1}{A small cat sitting on a chair by a window.}\\ $\bullet$ \textcolor{sp2}{A cat sitting on top of a wooden chair next to a window.}\\ $\bullet$ \textcolor{sp3}{A kitten balancing on a chair to look outside a window.}\\ $\bullet$ \textcolor{sp4}{A cat climbs up the back of  a chair to look out a window.}\\ $\bullet$ \textcolor{sp5}{A cat that is standing on a chair looking out of a window.}\end{tabular} &
  \begin{tabular}[c]{p{0.2\textwidth}}$\bullet$ \textcolor{sp1}{a white cat sitting on top of a window sill}\\ $\bullet$ \textcolor{sp1}{a white cat sitting on top of a chair}\\ $\bullet$ \textcolor{sp2}{a white cat sitting on top of a wooden chair looking out a window}\\ $\bullet$ \textcolor{sp3}{a cat sitting on a chair looking out the window}\\ $\bullet$ \textcolor{sp4}{a white cat sitting on top of a window sill}\end{tabular} &
  \begin{tabular}[c]{p{0.2\textwidth}}$\bullet$ \textcolor{sp1}{a cat perched on top of a table watching a lake through mesh}\\ $\bullet$ \textcolor{sp2}{a cat standing on an office chair looking out a window}\\ $\bullet$ \textcolor{sp3}{an animal sits on a chair staring out a window}\\ $\bullet$ \textcolor{sp4}{a kitten sitting alone by a window}\\ $\bullet$ \textcolor{sp5}{a UNK is looking out the window and cans are shining}\end{tabular} &
  \begin{tabular}[c]{p{0.2\textwidth}}$\bullet$ \textcolor{sp1}{a white cat sitting on top of a window}\\ $\bullet$ \textcolor{sp1}{a white cat sitting on top of a window}\\ $\bullet$ \textcolor{sp1}{a white cat sitting on top of a window}\\ $\bullet$ \textcolor{sp1}{a white cat sitting on top of a window}\\ $\bullet$ \textcolor{sp2}{a white kitty sitting on top of a window}\end{tabular}\\ \midrule
 &
   &
  \tiny CIDEr: 1.95, self-CIDEr: 0.55, TRP: 0.42 &
  \tiny CIDEr: 0.91, self-CIDEr: 0.94, TRP: 0.05 &
  \tiny CIDEr: 1.56, self-CIDEr: 0.24, TRP: 0.02 \\ \midrule
\multicolumn{1}{c}{\begin{minipage}[b]{0.6\columnwidth} \centering \raisebox{-.5\height}{\includegraphics[width=0.6\linewidth]{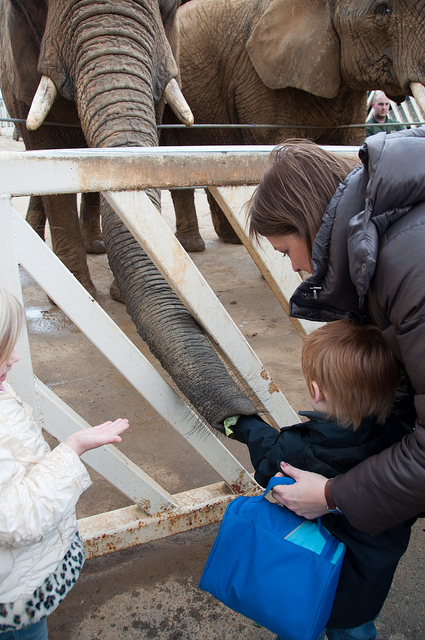}} \end{minipage}} &
  \begin{tabular}[c]{p{0.2\textwidth}}$\bullet$ \textcolor{sp1}{A mother and child standing in front of a group of elephants.}\\ $\bullet$ \textcolor{sp2}{A small boy is feeding the elephants through a fence.}\\ $\bullet$ \textcolor{sp3}{There is a woman and a boy feeding an elephant.}\\ $\bullet$ \textcolor{sp4}{An adult and child feeding an elephant in a contained area.  }\\ $\bullet$ \textcolor{sp5}{a woman and two children are feeding some elephants}\end{tabular} &
  \begin{tabular}[c]{p{0.2\textwidth}}$\bullet$ \textcolor{sp1}{a group of people looking at elephants}\\ $\bullet$ \textcolor{sp1}{a group of people looking at elephants in a fence}\\ $\bullet$ \textcolor{sp1}{a group of people feeding elephants at a fence}\\ $\bullet$ \textcolor{sp2}{a group of people feeding elephants at a fence}\\ $\bullet$ \textcolor{sp3}{a group of people looking at elephants in a fence}\end{tabular} &
  \begin{tabular}[c]{p{0.2\textwidth}}$\bullet$ \textcolor{sp1}{a close up of a person letting cut a small child that is looking at an}\\ $\bullet$ \textcolor{sp2}{a group of people feeding elephants on a stick}\\ $\bullet$ \textcolor{sp3}{a person leaning over a rail watching elephants at the zoo}\\ $\bullet$ \textcolor{sp4}{adult feeding baby elephant by man in standing next to woman}\\ $\bullet$ \textcolor{sp5}{a little boy is touching the trunk of an elephant at a zoo}\end{tabular} &
  \begin{tabular}[c]{p{0.2\textwidth}}$\bullet$ \textcolor{sp1}{a woman and a child looking at elephants}\\ $\bullet$ \textcolor{sp1}{a woman and a child looking at elephants}\\ $\bullet$ \textcolor{sp1}{a woman and a child looking at elephants}\\ $\bullet$ \textcolor{sp2}{a woman and a child looking at a herd of elephants}\\ $\bullet$ \textcolor{sp3}{a woman and a child looking at large elephants}\end{tabular}\\ \midrule
 &
   &
  \tiny CIDEr: 0.89, self-CIDEr: 0.45, TRP: -0.24 &
  \tiny CIDEr: 0.54, self-CIDEr: 0.97, TRP: -0.15 &
  \tiny CIDEr: 1.02, self-CIDEr: 0.31, TRP: -0.24 \\ \midrule
\multicolumn{1}{c}{\begin{minipage}[b]{0.6\columnwidth} \centering \raisebox{-.5\height}{\includegraphics[width=\linewidth]{samples/COCO_val2014_000000340332.jpg}} \end{minipage}} &
  \begin{tabular}[c]{p{0.2\textwidth}}$\bullet$ \textcolor{sp1}{An old opened trunk in an abandoned house}\\ $\bullet$ \textcolor{sp2}{an open chest in a run down apartment}\\ $\bullet$ \textcolor{sp3}{This is an image of a trunk in a damaged home.}\\ $\bullet$ \textcolor{sp4}{Large, open, empty chest in an abandoned, rundown buliding }\\ $\bullet$ \textcolor{sp5}{An old open trunk sits in a dilapidated house.}\end{tabular} &
  \begin{tabular}[c]{p{0.2\textwidth}}$\bullet$ \textcolor{sp1}{a broken suitcase sitting on the floor in a room}\\ $\bullet$ \textcolor{sp1}{a broken suitcase sitting on the floor in a room}\\ $\bullet$ \textcolor{sp2}{a broken suitcase sitting in the middle of a street}\\ $\bullet$ \textcolor{sp3}{a dirty old suitcase sitting on the ground}\\ $\bullet$ \textcolor{sp4}{an old old broken suitcase sitting in the street}\end{tabular} &
  \begin{tabular}[c]{p{0.2\textwidth}}$\bullet$ \textcolor{sp1}{a dirty stove that is in a snowy lot}\\ $\bullet$ \textcolor{sp2}{this is a broken looking under renovation in an abandoned kitchen}\\ $\bullet$ \textcolor{sp3}{a wooden box that has been destroyed with a broken stove and the bottom that was}\\ $\bullet$ \textcolor{sp4}{an old refrigerator}\\ $\bullet$ \textcolor{sp5}{the sink in the house have been turned down with the dead floor and is destroyed}\end{tabular} &
  \begin{tabular}[c]{p{0.2\textwidth}}$\bullet$ \textcolor{sp1}{an old suitcase sitting in the floor in a room}\\ $\bullet$ \textcolor{sp2}{an old suitcase sitting in the street in a room}\\ $\bullet$ \textcolor{sp3}{an old broken stove in a room with a wall}\\ $\bullet$ \textcolor{sp4}{an old broken old stove in a room}\\ $\bullet$ \textcolor{sp5}{an old stove sitting in a room in a house}\end{tabular}\\ \midrule
 &
   &
  \tiny CIDEr: 0.06, self-CIDEr: 0.57, TRP: -0.65 &
  \tiny CIDEr: 0.15, self-CIDEr: 0.95, TRP: -0.38 &
  \tiny CIDEr: 0.24, self-CIDEr: 0.76, TRP: -0.43 \\ \midrule
\multicolumn{1}{c}{\begin{minipage}[b]{0.6\columnwidth} \centering \raisebox{-.5\height}{\includegraphics[width=0.65\linewidth]{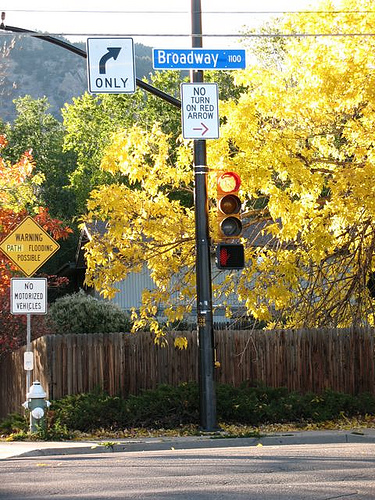}} \end{minipage}} &
  \begin{tabular}[c]{p{0.2\textwidth}}$\bullet$ \textcolor{sp1}{a green sign on a pole with a street light}\\ $\bullet$ \textcolor{sp2}{Street signage directs traffic moving right along Broadway.  }\\ $\bullet$ \textcolor{sp3}{a pole that has a sign on it}\\ $\bullet$ \textcolor{sp4}{A street with a stoplight on the yellow light. }\\ $\bullet$ \textcolor{sp5}{A picture of a traffic sign that says no turn on red arrow.}\end{tabular} &
  \begin{tabular}[c]{p{0.2\textwidth}}$\bullet$ \textcolor{sp1}{a traffic light and a street sign on a pole}\\ $\bullet$ \textcolor{sp1}{a street sign and a traffic light on a street}\\ $\bullet$ \textcolor{sp2}{a couple of street signs and a traffic light}\\ $\bullet$ \textcolor{sp3}{a traffic light and a street sign on a pole}\\ $\bullet$ \textcolor{sp4}{a couple of street signs on top of a street}\end{tabular} &
  \begin{tabular}[c]{p{0.2\textwidth}}$\bullet$ \textcolor{sp1}{a lamppost and street sign on the side of the road}\\ $\bullet$ \textcolor{sp2}{the traffic signal is at a city intersection near a caution light}\\ $\bullet$ \textcolor{sp3}{urban street intersection in UNK area with UNK gate sign and lantern}\\ $\bullet$ \textcolor{sp4}{a four way street intersection next to a fire hydrant and two street signs}\\ $\bullet$ \textcolor{sp5}{this is an image of an outdoor area}\end{tabular} &
  \begin{tabular}[c]{p{0.2\textwidth}}$\bullet$ \textcolor{sp1}{a street sign and a traffic light on a pole}\\ $\bullet$ \textcolor{sp2}{a traffic light and street signs on a pole}\\ $\bullet$ \textcolor{sp3}{a street sign and a traffic light on a street}\\ $\bullet$ \textcolor{sp4}{a red traffic light and a street sign on a pole}\\ $\bullet$ \textcolor{sp5}{a street sign and a traffic light on a road}\end{tabular}\\ \midrule
 &
   &
  \tiny CIDEr: 0.89, self-CIDEr: 0.63, TRP: -0.14 &
  \tiny CIDEr: 0.15, self-CIDEr: 0.98, TRP: -0.36 &
  \tiny CIDEr: 1.02, self-CIDEr: 0.56, TRP: -0.11 \\ \bottomrule
\end{tabular}%
}
\end{table*}

\begin{table*}[]
\centering
\scriptsize
\resizebox{\textwidth}{!}{%
\begin{tabular}{p{0.28\textwidth}<{\centering}p{0.21\textwidth}<{\centering}p{0.21\textwidth}<{\centering}p{0.21\textwidth}<{\centering}p{0.21\textwidth}<{\centering}}
\toprule
\normalsize Image &
\normalsize  Human annotations &
\normalsize  VaT(ours, msc) &
\normalsize  Diverse model &
\normalsize  Accurate model \\ \midrule
\multicolumn{1}{c}{\begin{minipage}[b]{0.6\columnwidth} \centering \raisebox{-.5\height}{\includegraphics[width=\linewidth]{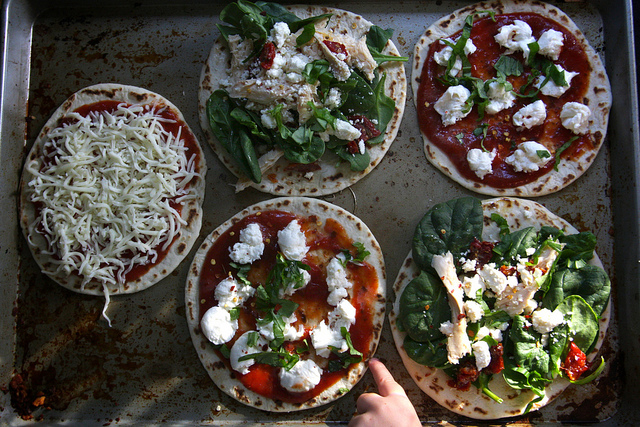}} \end{minipage}} &
  \begin{tabular}[c]{p{0.2\textwidth}}$\bullet$ \textcolor{sp1}{A set of five pizzas sitting next to each other each with different toppings.}\\ $\bullet$ \textcolor{sp2}{A handful of prepared pizzas sit next to each other.}\\ $\bullet$ \textcolor{sp3}{Five uncooked pizzas with a variety of different toppings.}\\ $\bullet$ \textcolor{sp4}{Five unbaked pizzas that include various types of cheeses.}\\ $\bullet$ \textcolor{sp5}{Five different pizzas are being prepared over a metal tray.}\end{tabular} &
  \begin{tabular}[c]{p{0.2\textwidth}}$\bullet$ \textcolor{sp1}{a group of pizzas sitting on top of a table}\\ $\bullet$ \textcolor{sp1}{a group of pizzas sitting on top of a table}\\ $\bullet$ \textcolor{sp1}{four pizzas sitting on top of a table}\\ $\bullet$ \textcolor{sp2}{a group of pizzas sitting on a table}\\ $\bullet$ \textcolor{sp3}{a group of pizzas sitting on top of a table}\end{tabular} &
  \begin{tabular}[c]{p{0.2\textwidth}}$\bullet$ \textcolor{sp1}{four pizzas with different toppings on a buffet line}\\ $\bullet$ \textcolor{sp2}{a set of six small pizzas with green leaves on them}\\ $\bullet$ \textcolor{sp3}{five uncooked pizzas with food are on wooden platters}\\ $\bullet$ \textcolor{sp4}{several uncooked pizzas in different shapes and sizes and styles}\\ $\bullet$ \textcolor{sp5}{six uncooked pizzas on display with puffs on them}\end{tabular} &
  \begin{tabular}[c]{p{0.2\textwidth}}$\bullet$ \textcolor{sp1}{a group of pizzas sitting on top of a table}\\ $\bullet$ \textcolor{sp1}{a group of pizzas sitting on top of a table}\\ $\bullet$ \textcolor{sp1}{a group of pizzas sitting on top of a table}\\ $\bullet$ \textcolor{sp1}{a group of pizzas sitting on top of a table}\\ $\bullet$ \textcolor{sp2}{a group of pizzas are sitting on a table}\end{tabular}\\ \midrule
 &
   &
  \tiny CIDEr: 0.95, self-CIDEr: 0.29, TRP: -0.29 &
  \tiny CIDEr: 0.87, self-CIDEr: 0.90, TRP: 0.01 &
  \tiny CIDEr: 1.00, self-CIDEr: 0.21, TRP: -0.32 \\ \midrule
\multicolumn{1}{c}{\begin{minipage}[b]{0.6\columnwidth} \centering \raisebox{-.5\height}{\includegraphics[width=0.6\linewidth]{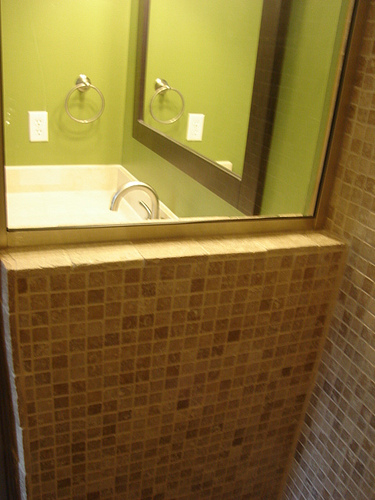}} \end{minipage}} &
  \begin{tabular}[c]{p{0.2\textwidth}}$\bullet$ \textcolor{sp1}{a small bathroom with a mirror and sink}\\ $\bullet$ \textcolor{sp2}{A tile wall in a bathroom below a mirror.}\\ $\bullet$ \textcolor{sp3}{A tiled mosaic empty shower stall with bathroom mirror.}\\ $\bullet$ \textcolor{sp4}{The mirror is showing the reflection of the bathroom sink}\\ $\bullet$ \textcolor{sp5}{A sink with tile on the wall near it and green paint on the wall}\end{tabular} &
  \begin{tabular}[c]{p{0.2\textwidth}}$\bullet$ \textcolor{sp1}{a bathroom with a sink and a mirror}\\ $\bullet$ \textcolor{sp1}{a bathroom with a sink and a mirror}\\ $\bullet$ \textcolor{sp1}{a bathroom with a sink and a mirror}\\ $\bullet$ \textcolor{sp1}{a bathroom with a sink and a mirror}\\ $\bullet$ \textcolor{sp2}{a bathroom with a mirror and a sink}\end{tabular} &
  \begin{tabular}[c]{p{0.2\textwidth}}$\bullet$ \textcolor{sp1}{a bathroom has a sink and a mirror in it}\\ $\bullet$ \textcolor{sp2}{a brown marble tiled bathroom with sink mirror bathtub and towel rack}\\ $\bullet$ \textcolor{sp3}{a tiled bathroom with two sinks under a large mirror}\\ $\bullet$ \textcolor{sp4}{a bathroom sink and mirror in a bathroom}\\ $\bullet$ \textcolor{sp5}{a tiled wall with a basin sink is shown here with no faucet in hand}\end{tabular} &
  \begin{tabular}[c]{p{0.2\textwidth}}$\bullet$ \textcolor{sp1}{a bathroom with a sink and a mirror}\\ $\bullet$ \textcolor{sp1}{a bathroom with a sink and a mirror}\\ $\bullet$ \textcolor{sp1}{a bathroom with a sink and a mirror}\\ $\bullet$ \textcolor{sp1}{a bathroom with a sink and a mirror}\\ $\bullet$ \textcolor{sp1}{a bathroom with a sink and a mirror}\end{tabular}\\ \midrule
 &
   &
  \tiny CIDEr: 1.66, self-CIDEr: 0.21, TRP: 0.06 &
  \tiny CIDEr: 0.78, self-CIDEr: 0.87, TRP: -0.07 &
  \tiny CIDEr: 1.55, self-CIDEr: 0.00, TRP: -0.12 \\ \midrule
\multicolumn{1}{c}{\begin{minipage}[b]{0.6\columnwidth} \centering \raisebox{-.5\height}{\includegraphics[width=0.8\linewidth]{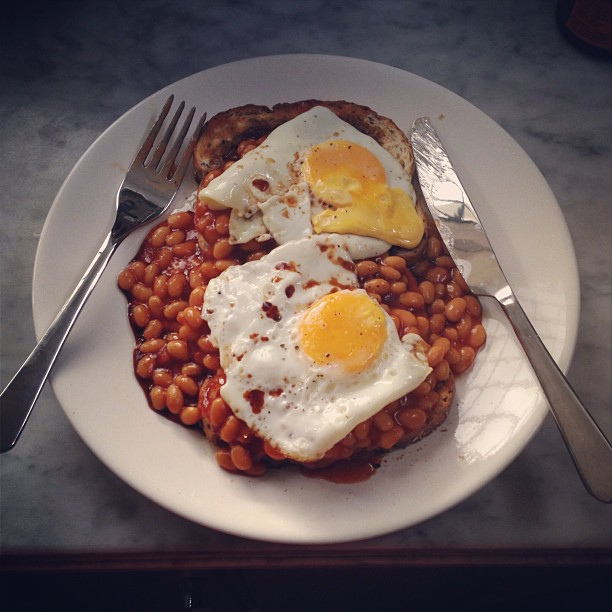}} \end{minipage}} &
  \begin{tabular}[c]{p{0.2\textwidth}}$\bullet$ \textcolor{sp1}{A plate of beans and eggs with a knife and fork on it.}\\ $\bullet$ \textcolor{sp2}{A plate of beans, eggs, and toast is on the table.}\\ $\bullet$ \textcolor{sp3}{A plate on a table containing beans, eggs, toast and silverware.}\\ $\bullet$ \textcolor{sp4}{A plate of breakfast food is sitting on the table.}\\ $\bullet$ \textcolor{sp5}{There are eggs and beans on a plate.}\end{tabular} &
  \begin{tabular}[c]{p{0.2\textwidth}}$\bullet$ \textcolor{sp1}{a plate of breakfast food with eggs and beans}\\ $\bullet$ \textcolor{sp2}{a plate of breakfast food with eggs and a fork}\\ $\bullet$ \textcolor{sp3}{a plate of food with eggs and eggs on a table}\\ $\bullet$ \textcolor{sp4}{a plate of food with eggs and beans on a table}\\ $\bullet$ \textcolor{sp5}{a breakfast plate with eggs and beans on a table}\end{tabular} &
  \begin{tabular}[c]{p{0.2\textwidth}}$\bullet$ \textcolor{sp1}{plate with grits and eggs with eggs on a black table}\\ $\bullet$ \textcolor{sp2}{two eggs on a plate with mac and cheese next to a fork and a spoon}\\ $\bullet$ \textcolor{sp3}{the breakfast has eggs eggs and toast on it}\\ $\bullet$ \textcolor{sp4}{breakfast of eggs beans and a motor fathers breakfast}\\ $\bullet$ \textcolor{sp5}{there are some foods on this white dish}\end{tabular} &
  \begin{tabular}[c]{p{0.2\textwidth}}$\bullet$ \textcolor{sp1}{a plate of food with eggs and beans on a table}\\ $\bullet$ \textcolor{sp1}{a plate of food with eggs and beans on a table}\\ $\bullet$ \textcolor{sp1}{a plate of food with eggs and beans on a table}\\ $\bullet$ \textcolor{sp1}{a plate of food with eggs and beans on a table}\\ $\bullet$ \textcolor{sp2}{a plate of food with egg and blue beans on a table}\end{tabular}\\ \midrule
 &
   &
  \tiny CIDEr: 2.10, self-CIDEr: 0.58, TRP: 0.52 &
  \tiny CIDEr: 0.64, self-CIDEr: 0.91, TRP: -0.12 &
  \tiny CIDEr: 2.28, self-CIDEr: 0.23, TRP: 0.43 \\ \midrule
\multicolumn{1}{c}{\begin{minipage}[b]{0.6\columnwidth} \centering \raisebox{-.5\height}{\includegraphics[width=0.6\linewidth]{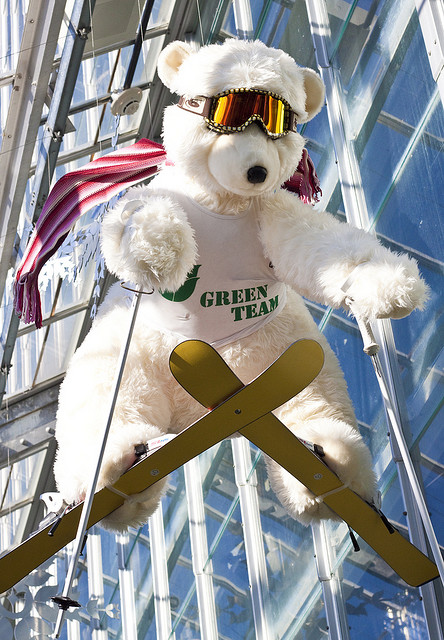}} \end{minipage}} &
  \begin{tabular}[c]{p{0.2\textwidth}}$\bullet$ \textcolor{sp1}{A large teddybear float is on snow skis.}\\ $\bullet$ \textcolor{sp2}{A white teddy bear posed on skis with ski poles.}\\ $\bullet$ \textcolor{sp3}{a stuffed bear with goggled on wearing snow skies }\\ $\bullet$ \textcolor{sp4}{A stuffed polar bear has goggles and skis.}\\ $\bullet$ \textcolor{sp5}{There is a giant stuffed bear wearing skis}\end{tabular} &
  \begin{tabular}[c]{p{0.2\textwidth}}$\bullet$ \textcolor{sp1}{a dog wearing a hat sitting on a boat}\\ $\bullet$ \textcolor{sp2}{a dressed in dog is sitting on a boat}\\ $\bullet$ \textcolor{sp3}{a teddy bear wearing a hat sitting on a boat}\\ $\bullet$ \textcolor{sp4}{a teddy bear wearing a headband sitting on a boat}\\ $\bullet$ \textcolor{sp5}{a dog in a santa suit sitting on a boat}\end{tabular} &
  \begin{tabular}[c]{p{0.2\textwidth}}$\bullet$ \textcolor{sp1}{a UNK teddy bear laying on a white object}\\ $\bullet$ \textcolor{sp2}{a dog is sitting in a cat outfit while sitting in a boat}\\ $\bullet$ \textcolor{sp3}{a teddy bear dressed as a lobster in uniform}\\ $\bullet$ \textcolor{sp4}{a white teddy bear sitting by a fence}\\ $\bullet$ \textcolor{sp5}{a yellow teddy bear dressed in german clothing}\end{tabular} &
  \begin{tabular}[c]{p{0.2\textwidth}}$\bullet$ \textcolor{sp1}{a teddy bear wearing a hat sitting on a ski lift}\\ $\bullet$ \textcolor{sp1}{a teddy bear wearing a hat sitting on a ski lift}\\ $\bullet$ \textcolor{sp2}{a stuffed teddy bear wearing a hat sitting on a ski lift}\\ $\bullet$ \textcolor{sp3}{a teddy bear sitting on top of a UNK}\\ $\bullet$ \textcolor{sp4}{a stuffed teddy bear wearing a hat sitting on a UNK}\end{tabular}\\ \midrule
 &
   &
  \tiny CIDEr: 0.25, self-CIDEr: 0.75, TRP: -0.44 &
  \tiny CIDEr: 0.35, self-CIDEr: 0.91, TRP: -0.29 &
  \tiny CIDEr: 0.58, self-CIDEr: 0.47, TRP: -0.41 \\ \midrule
\multicolumn{1}{c}{\begin{minipage}[b]{0.6\columnwidth} \centering \raisebox{-.5\height}{\includegraphics[width=0.6\linewidth]{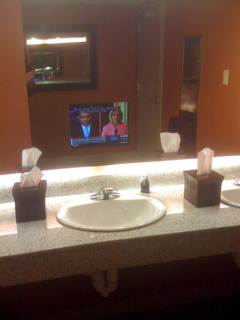}} \end{minipage}} &
  \begin{tabular}[c]{p{0.2\textwidth}}$\bullet$ \textcolor{sp1}{A bathroom with a reflection of a television and a sink.}\\ $\bullet$ \textcolor{sp2}{A bathroom with a sink and mirror and tissues on the counter.}\\ $\bullet$ \textcolor{sp3}{Bathroom area with multiple sinks and mirrors with television reflected.}\\ $\bullet$ \textcolor{sp4}{A sink and mirror in a motel room.}\\ $\bullet$ \textcolor{sp5}{A bathroom with a television, sink and two boxes of tissues. }\end{tabular} &
  \begin{tabular}[c]{p{0.2\textwidth}}$\bullet$ \textcolor{sp1}{a bathroom with a sink and a mirror}\\ $\bullet$ \textcolor{sp1}{a bathroom with a sink and a mirror}\\ $\bullet$ \textcolor{sp1}{a bathroom with a sink and a mirror}\\ $\bullet$ \textcolor{sp2}{a hotel bathroom with a sink and a tv}\\ $\bullet$ \textcolor{sp3}{a bathroom sink and a flat screen tv in a bathroom}\end{tabular} &
  \begin{tabular}[c]{p{0.2\textwidth}}$\bullet$ \textcolor{sp1}{a bathroom has a large double sink a large framed picture and two square candles}\\ $\bullet$ \textcolor{sp2}{a bathroom sink with two candles in it}\\ $\bullet$ \textcolor{sp3}{a bathroom vanity is photographed with two photographers displayed}\\ $\bullet$ \textcolor{sp4}{two large sinks near two mirrors in a public restroom}\\ $\bullet$ \textcolor{sp5}{this picture is of someone singing at his bathroom belongings}\end{tabular} &
  \begin{tabular}[c]{p{0.2\textwidth}}$\bullet$ \textcolor{sp1}{a bathroom with a sink and a mirror}\\ $\bullet$ \textcolor{sp1}{a bathroom with a sink and a mirror}\\ $\bullet$ \textcolor{sp1}{a bathroom with a sink and a mirror}\\ $\bullet$ \textcolor{sp1}{a bathroom with a sink and a mirror}\\ $\bullet$ \textcolor{sp2}{a bathroom with a sink and a picture on the mirror}\end{tabular}\\ \midrule
 &
   &
  \tiny CIDEr: 1.49, self-CIDEr: 0.43, TRP: 0.09 &
  \tiny CIDEr: 0.30, self-CIDEr: 0.95, TRP: -0.30 &
  \tiny CIDEr: 1.87, self-CIDEr: 0.19, TRP: 0.17 \\ \midrule
\multicolumn{1}{c}{\begin{minipage}[b]{0.6\columnwidth} \centering \raisebox{-.5\height}{\includegraphics[width=\linewidth]{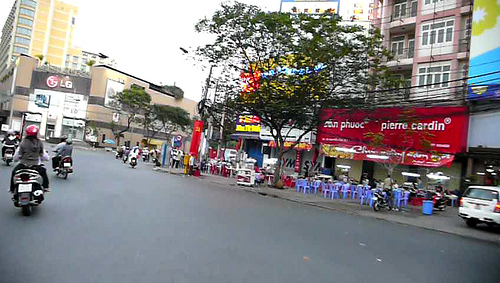}} \end{minipage}} &
  \begin{tabular}[c]{p{0.2\textwidth}}$\bullet$ \textcolor{sp1}{A couple of people that are sitting on mopeds.}\\ $\bullet$ \textcolor{sp2}{a street that has some motorcycles going down it}\\ $\bullet$ \textcolor{sp3}{A city street with multiple shops and people riding motorcycles.}\\ $\bullet$ \textcolor{sp4}{People on motorbikes driving down the road in a city.}\\ $\bullet$ \textcolor{sp5}{A few people on motor scooters riding down a street.}\end{tabular} &
  \begin{tabular}[c]{p{0.2\textwidth}}$\bullet$ \textcolor{sp1}{a group of people riding motorcycles down a city street}\\ $\bullet$ \textcolor{sp1}{a group of people riding motorcycles down a city street}\\ $\bullet$ \textcolor{sp1}{a group of people riding motorcycles down a street}\\ $\bullet$ \textcolor{sp1}{a group of people riding motorcycles down a city street}\\ $\bullet$ \textcolor{sp2}{a group of people riding motorcycles down a street}\end{tabular} &
  \begin{tabular}[c]{p{0.2\textwidth}}$\bullet$ \textcolor{sp1}{motorcycle riders are at an intersection in a city}\\ $\bullet$ \textcolor{sp2}{people are riding their bikes down a street}\\ $\bullet$ \textcolor{sp3}{suv and riders are going in bicycles down a city street}\\ $\bullet$ \textcolor{sp4}{many motorcycle riders ride down the city street}\\ $\bullet$ \textcolor{sp5}{an foreign country bikers parade along this busy city street}\end{tabular} &
  \begin{tabular}[c]{p{0.2\textwidth}}$\bullet$ \textcolor{sp1}{a group of people riding motorcycles down a city street}\\ $\bullet$ \textcolor{sp1}{a group of people riding motorcycles down a city street}\\ $\bullet$ \textcolor{sp1}{a group of people riding motorcycles down a city street}\\ $\bullet$ \textcolor{sp1}{a group of people riding motorcycles down a city street}\\ $\bullet$ \textcolor{sp2}{a group of people on motorcycles on a city street}\end{tabular}\\ \midrule
 &
   &
  \tiny CIDEr: 1.18, self-CIDEr: 0.15, TRP: -0.24 &
  \tiny CIDEr: 0.28, self-CIDEr: 0.91, TRP: -0.33 &
  \tiny CIDEr: 1.15, self-CIDEr: 0.23, TRP: -0.21 \\ \bottomrule
\end{tabular}%
}
\end{table*}

\begin{table*}[]
\centering
\scriptsize
\resizebox{\textwidth}{!}{%
\begin{tabular}{p{0.28\textwidth}<{\centering}p{0.21\textwidth}<{\centering}p{0.21\textwidth}<{\centering}p{0.21\textwidth}<{\centering}p{0.21\textwidth}<{\centering}}
\toprule
\normalsize Image &
\normalsize  Human annotations &
\normalsize  VaT(ours, msc) &
\normalsize  Diverse model &
\normalsize  Accurate model \\ \midrule
\multicolumn{1}{c}{\begin{minipage}[b]{0.6\columnwidth} \centering \raisebox{-.5\height}{\includegraphics[width=\linewidth]{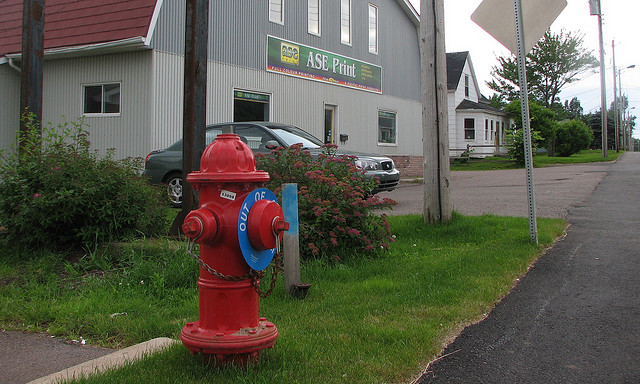}} \end{minipage}} &
  \begin{tabular}[c]{p{0.2\textwidth}}$\bullet$ \textcolor{sp1}{Red fire hydrant on grass in front of a black street.}\\ $\bullet$ \textcolor{sp2}{A fire hydrant on the side of the road outside a business.}\\ $\bullet$ \textcolor{sp3}{A fire hydrant with an out of order sign on a street side.}\\ $\bullet$ \textcolor{sp4}{A red fire hydrant in grassy area next to street.}\\ $\bullet$ \textcolor{sp5}{A fire hydrant that is in the grass.}\end{tabular} &
  \begin{tabular}[c]{p{0.2\textwidth}}$\bullet$ \textcolor{sp1}{a red fire hydrant in the grass next to a street}\\ $\bullet$ \textcolor{sp1}{a red fire hydrant in the side of a street}\\ $\bullet$ \textcolor{sp1}{a red fire hydrant on the side of a street}\\ $\bullet$ \textcolor{sp2}{a red fire hydrant on the side of a street}\\ $\bullet$ \textcolor{sp3}{a red fire hydrant on the side of a street}\end{tabular} &
  \begin{tabular}[c]{p{0.2\textwidth}}$\bullet$ \textcolor{sp1}{a fire hydrant on grass by the street}\\ $\bullet$ \textcolor{sp2}{a red fire hydrant is in a very popular location}\\ $\bullet$ \textcolor{sp3}{a red fire hydrant on the street in front of a brick building}\\ $\bullet$ \textcolor{sp4}{a red fire hydrant in grassy area next to bushes and trees}\\ $\bullet$ \textcolor{sp5}{the street is full of cars and a fire hydrant sit in the grass beside a}\end{tabular} &
  \begin{tabular}[c]{p{0.2\textwidth}}$\bullet$ \textcolor{sp1}{a red fire hydrant on the side of a street}\\ $\bullet$ \textcolor{sp1}{a red fire hydrant on the side of a street}\\ $\bullet$ \textcolor{sp1}{a red fire hydrant on the side of a street}\\ $\bullet$ \textcolor{sp1}{a red fire hydrant on the side of a street}\\ $\bullet$ \textcolor{sp2}{a red fire hydrant sitting on the side of a street}\end{tabular}\\ \midrule
 &
   &
  \tiny CIDEr: 2.95, self-CIDEr: 0.36, TRP: 0.88 &
  \tiny CIDEr: 1.95, self-CIDEr: 0.82, TRP: 0.58 &
  \tiny CIDEr: 3.01, self-CIDEr: 0.18, TRP: 0.81 \\ \midrule
\multicolumn{1}{c}{\begin{minipage}[b]{0.6\columnwidth} \centering \raisebox{-.5\height}{\includegraphics[width=\linewidth]{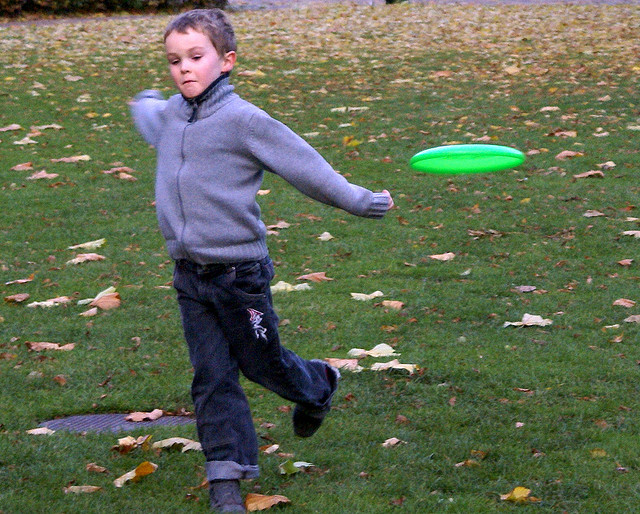}} \end{minipage}} &
  \begin{tabular}[c]{p{0.2\textwidth}}$\bullet$ \textcolor{sp1}{A young boy standing next to a green frisbee.}\\ $\bullet$ \textcolor{sp2}{a young boy in a gray sweater throwing a green Frisbee in a park}\\ $\bullet$ \textcolor{sp3}{a llittle boy in a sweater is throwing a frisbee}\\ $\bullet$ \textcolor{sp4}{A young boy throwing a frisbee in a yard.}\\ $\bullet$ \textcolor{sp5}{A young bow throws a Frisbee behind him}\end{tabular} &
  \begin{tabular}[c]{p{0.2\textwidth}}$\bullet$ \textcolor{sp1}{a young man throwing a frisbee in a field}\\ $\bullet$ \textcolor{sp2}{a young man throwing a green frisbee in a field}\\ $\bullet$ \textcolor{sp3}{a young boy throwing a frisbee in a field}\\ $\bullet$ \textcolor{sp4}{a young boy throwing a blue frisbee in a field}\\ $\bullet$ \textcolor{sp5}{a young boy throwing a green frisbee in the park}\end{tabular} &
  \begin{tabular}[c]{p{0.2\textwidth}}$\bullet$ \textcolor{sp1}{a young man tossing a frisbee in a grassy field}\\ $\bullet$ \textcolor{sp2}{a young boy looking at a frisbee on a green lawn}\\ $\bullet$ \textcolor{sp3}{an older gentleman tossing a green disk to catch it}\\ $\bullet$ \textcolor{sp4}{a guy throwing a green flying disc in a grassy area}\\ $\bullet$ \textcolor{sp5}{a young boy has thrown a frisbee out on a grassy area}\end{tabular} &
  \begin{tabular}[c]{p{0.2\textwidth}}$\bullet$ \textcolor{sp1}{a young boy throwing a frisbee in the grass}\\ $\bullet$ \textcolor{sp1}{a young boy throwing a frisbee in the grass}\\ $\bullet$ \textcolor{sp1}{a young boy throwing a frisbee in the grass}\\ $\bullet$ \textcolor{sp2}{a young boy throwing a frisbee in a field}\\ $\bullet$ \textcolor{sp3}{a young boy playing with a frisbee in a field}\end{tabular}\\ \midrule
 &
   &
  \tiny CIDEr: 2.53, self-CIDEr: 0.59, TRP: 0.77 &
  \tiny CIDEr: 0.77, self-CIDEr: 0.92, TRP: -0.04 &
  \tiny CIDEr: 2.73, self-CIDEr: 0.33, TRP: 0.74 \\ \midrule
\multicolumn{1}{c}{\begin{minipage}[b]{0.6\columnwidth} \centering \raisebox{-.5\height}{\includegraphics[width=0.6\linewidth]{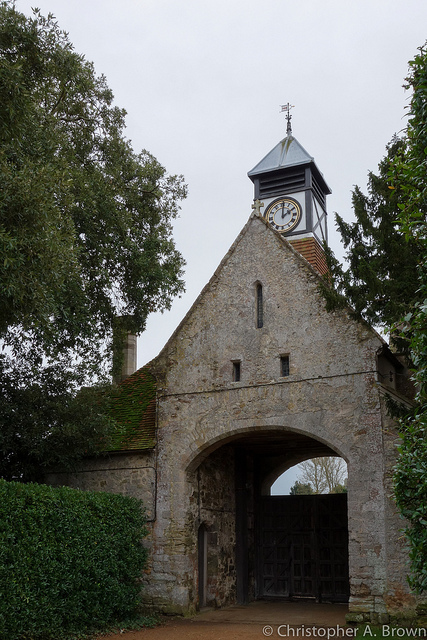}} \end{minipage}} &
  \begin{tabular}[c]{p{0.2\textwidth}}$\bullet$ \textcolor{sp1}{A large clock tower over a church next to trees.}\\ $\bullet$ \textcolor{sp2}{A large clock tower above a masonry building.}\\ $\bullet$ \textcolor{sp3}{The building has an archway going through it.}\\ $\bullet$ \textcolor{sp4}{The entrance to the old chapel is waiting for visitors,}\\ $\bullet$ \textcolor{sp5}{A stone entryway that is topped with a clock tower }\end{tabular} &
  \begin{tabular}[c]{p{0.2\textwidth}}$\bullet$ \textcolor{sp1}{a building with a clock tower on top of it}\\ $\bullet$ \textcolor{sp1}{a building with a clock tower on top of it}\\ $\bullet$ \textcolor{sp1}{a large building with a clock tower on top of it}\\ $\bullet$ \textcolor{sp2}{a church with a clock tower on the top of it}\\ $\bullet$ \textcolor{sp3}{a building with a clock tower on top of it}\end{tabular} &
  \begin{tabular}[c]{p{0.2\textwidth}}$\bullet$ \textcolor{sp1}{an old structure with a casting iron and a clock on a tower}\\ $\bullet$ \textcolor{sp2}{a clock tower with a bell on it near trees}\\ $\bullet$ \textcolor{sp3}{an older stone tower has a clock at the top}\\ $\bullet$ \textcolor{sp4}{copper stone clock tower is above the trees}\\ $\bullet$ \textcolor{sp5}{there is a big brick tower with a clock in the middle}\end{tabular} &
  \begin{tabular}[c]{p{0.2\textwidth}}$\bullet$ \textcolor{sp1}{a church with a clock tower on top of it}\\ $\bullet$ \textcolor{sp1}{a church with a clock tower on top of it}\\ $\bullet$ \textcolor{sp1}{a church with a clock tower on top of it}\\ $\bullet$ \textcolor{sp1}{a church with a clock tower on top of it}\\ $\bullet$ \textcolor{sp2}{a building with a clock tower on top of it}\end{tabular}\\ \midrule
 &
   &
  \tiny CIDEr: 1.09, self-CIDEr: 0.31, TRP: -0.20 &
  \tiny CIDEr: 0.55, self-CIDEr: 0.90, TRP: -0.18 &
  \tiny CIDEr: 1.11, self-CIDEr: 0.17, TRP: -0.27 \\ \midrule
\multicolumn{1}{c}{\begin{minipage}[b]{0.6\columnwidth} \centering \raisebox{-.5\height}{\includegraphics[width=0.6\linewidth]{samples/COCO_val2014_000000007088.jpg}} \end{minipage}} &
  \begin{tabular}[c]{p{0.2\textwidth}}$\bullet$ \textcolor{sp1}{A little girl is holding an umbrella on a wet day.}\\ $\bullet$ \textcolor{sp2}{A small child stands in the rain with an umbrella.}\\ $\bullet$ \textcolor{sp3}{A perfect pose by a little girl with an umbrella in the rain.}\\ $\bullet$ \textcolor{sp4}{A little girl holding an umbrella in the rain. }\\ $\bullet$ \textcolor{sp5}{A little girl holding an umbrella while standing outside.}\end{tabular} &
  \begin{tabular}[c]{p{0.2\textwidth}}$\bullet$ \textcolor{sp1}{a little girl standing in the rain with an umbrella}\\ $\bullet$ \textcolor{sp1}{a little girl holding an umbrella in the rain}\\ $\bullet$ \textcolor{sp1}{a little girl holding an umbrella in the rain}\\ $\bullet$ \textcolor{sp2}{a little girl holding an umbrella in the rain}\\ $\bullet$ \textcolor{sp3}{a little girl holding an umbrella on a street}\end{tabular} &
  \begin{tabular}[c]{p{0.2\textwidth}}$\bullet$ \textcolor{sp1}{a young child standing in the rain near a bush holding an umbrella}\\ $\bullet$ \textcolor{sp2}{a young girl holding an umbrella standing in the rain}\\ $\bullet$ \textcolor{sp3}{small girl holding an umbrella in driveway outside near flowers}\\ $\bullet$ \textcolor{sp4}{girl standing in rain with an umbrella on top of her head}\\ $\bullet$ \textcolor{sp5}{a little girl with an umbrella is standing outside}\end{tabular} &
  \begin{tabular}[c]{p{0.2\textwidth}}$\bullet$ \textcolor{sp1}{a little girl holding an umbrella in the rain}\\ $\bullet$ \textcolor{sp1}{a little girl holding an umbrella in the rain}\\ $\bullet$ \textcolor{sp1}{a little girl holding an umbrella in the rain}\\ $\bullet$ \textcolor{sp2}{a little girl standing in the rain with an umbrella}\\ $\bullet$ \textcolor{sp3}{a young girl holding an umbrella in the rain}\end{tabular}\\ \midrule
 &
   &
  \tiny CIDEr: 4.32, self-CIDEr: 0.38, TRP: 1.67 &
  \tiny CIDEr: 2.10, self-CIDEr: 0.83, TRP: 0.67 &
  \tiny CIDEr: 4.27, self-CIDEr: 0.35, TRP: 1.63 \\ \midrule
\multicolumn{1}{c}{\begin{minipage}[b]{0.6\columnwidth} \centering \raisebox{-.5\height}{\includegraphics[width=\linewidth]{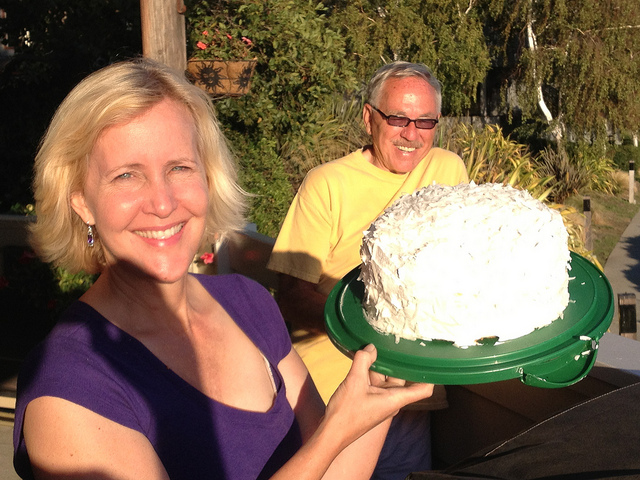}} \end{minipage}} &
  \begin{tabular}[c]{p{0.2\textwidth}}$\bullet$ \textcolor{sp1}{A woman is showing off a cake she baked while a man looks on. }\\ $\bullet$ \textcolor{sp2}{A lady wearing a purple shirt holding a cake.}\\ $\bullet$ \textcolor{sp3}{A man smiles in the background as a woman holds a cake. }\\ $\bullet$ \textcolor{sp4}{A woman standing outside holding a coconut cake with a man looking on.}\\ $\bullet$ \textcolor{sp5}{A woman smiles while holding up a cake.}\end{tabular} &
  \begin{tabular}[c]{p{0.2\textwidth}}$\bullet$ \textcolor{sp1}{a man and a woman holding a cake}\\ $\bullet$ \textcolor{sp1}{a man and a woman holding a cake}\\ $\bullet$ \textcolor{sp1}{a man and a woman holding a cake}\\ $\bullet$ \textcolor{sp2}{a man and a woman holding a green cake}\\ $\bullet$ \textcolor{sp3}{a woman holding a green cake next to a man}\end{tabular} &
  \begin{tabular}[c]{p{0.2\textwidth}}$\bullet$ \textcolor{sp1}{group of ladies sitting on a park bench in front of a cake}\\ $\bullet$ \textcolor{sp2}{a man and a woman smile as they hold a cake}\\ $\bullet$ \textcolor{sp3}{two women holding a plate with a cake on it smiling and a man holding a}\\ $\bullet$ \textcolor{sp4}{a girl with her arms covered cake and a cake on a table}\\ $\bullet$ \textcolor{sp5}{the woman and a man both have a cake in front of them}\end{tabular} &
  \begin{tabular}[c]{p{0.2\textwidth}}$\bullet$ \textcolor{sp1}{a man and a woman holding a cake}\\ $\bullet$ \textcolor{sp1}{a man and a woman holding a cake}\\ $\bullet$ \textcolor{sp2}{a man and a woman sitting in front of a birthday cake}\\ $\bullet$ \textcolor{sp3}{a man and a woman sitting holding a cake}\\ $\bullet$ \textcolor{sp4}{two people standing next to a table with a cake}\end{tabular}\\ \midrule
 &
   &
  \tiny CIDEr: 1.28, self-CIDEr: 0.36, TRP: -0.07 &
  \tiny CIDEr: 0.56, self-CIDEr: 0.91, TRP: -0.17 &
  \tiny CIDEr: 1.19, self-CIDEr: 0.59, TRP: 0.01 \\ \midrule
\multicolumn{1}{c}{\begin{minipage}[b]{0.6\columnwidth} \centering \raisebox{-.5\height}{\includegraphics[width=\linewidth]{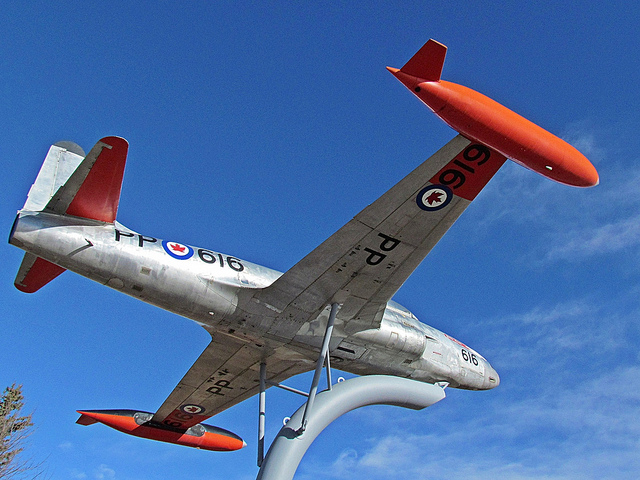}} \end{minipage}} &
  \begin{tabular}[c]{p{0.2\textwidth}}$\bullet$ \textcolor{sp1}{a air plane that is a model conected to a pole.}\\ $\bullet$ \textcolor{sp2}{An old missile carrier hangs in a glider}\\ $\bullet$ \textcolor{sp3}{A Canada Airforce plane mounted on a pole for display}\\ $\bullet$ \textcolor{sp4}{An old fighter jet mounted on display outside.}\\ $\bullet$ \textcolor{sp5}{An airplane mounted on a pole for display.}\end{tabular} &
  \begin{tabular}[c]{p{0.2\textwidth}}$\bullet$ \textcolor{sp1}{a red and white airplane flying in the sky}\\ $\bullet$ \textcolor{sp1}{a red and white airplane flying in the sky}\\ $\bullet$ \textcolor{sp1}{a red and blue airplane flying in the sky}\\ $\bullet$ \textcolor{sp2}{a large airplane flying in the blue sky}\\ $\bullet$ \textcolor{sp3}{a red and white airplane flying in the sky}\end{tabular} &
  \begin{tabular}[c]{p{0.2\textwidth}}$\bullet$ \textcolor{sp1}{a large older jet flying through a cloudy blue sky}\\ $\bullet$ \textcolor{sp2}{a mid sized air force plane flying overhead}\\ $\bullet$ \textcolor{sp3}{a small airplane flying high up in the sky}\\ $\bullet$ \textcolor{sp4}{the sleek top of a jet ladder}\\ $\bullet$ \textcolor{sp5}{a propeller plane flying through blue cloudy skies}\end{tabular} &
  \begin{tabular}[c]{p{0.2\textwidth}}$\bullet$ \textcolor{sp1}{a red and blue plane flying in the sky}\\ $\bullet$ \textcolor{sp1}{a red and blue plane flying in the sky}\\ $\bullet$ \textcolor{sp1}{a red and blue plane flying in the sky}\\ $\bullet$ \textcolor{sp1}{a red and blue plane flying in the sky}\\ $\bullet$ \textcolor{sp2}{a red and blue airplane flying in the sky}\end{tabular}\\ \midrule
 &
   &
  \tiny CIDEr: 0.13, self-CIDEr: 0.39, TRP: -0.71 &
  \tiny CIDEr: 0.10, self-CIDEr: 0.95, TRP: -0.41 &
  \tiny CIDEr: 0.15, self-CIDEr: 0.22, TRP: -0.79 \\ \midrule
\end{tabular}%
}
\end{table*}

\end{document}